\ifdefined\pdfminorversion\pdfminorversion=7\fi
\PassOptionsToPackage{table}{xcolor}
\PassOptionsToPackage{normalem}{ulem}
\documentclass[]{bytedance_seed}

\DeclareFontShape{T1}{bytesans}{b}{n}{<-> s * [1] ./seed/bytesans}{}
\DeclareFontShape{T1}{bytesans}{bx}{n}{<-> s * [1] ./seed/bytesans}{}
\hypersetup{
    pdftitle={Reinforcement Learning Done Right for Explicit Confidence Targets},
    pdfauthor={Zefeng He, Yichen He, Yuan Lin, Yang Yang, Yu Cheng}
}

\usepackage{amsmath,amsfonts,bm}

\def\eqref#1{equation~\ref{#1}}

\def\1{\bm{1}}

\DeclareMathAlphabet{\mathsfit}{\encodingdefault}{\sfdefault}{m}{sl}
\SetMathAlphabet{\mathsfit}{bold}{\encodingdefault}{\sfdefault}{bx}{n}

\usepackage{url}
\usepackage{wrapfig}
\usepackage{enumitem}

\titleformat*{\paragraph}{\normalfont\bfseries}

\patchcmd{\mymaketitle}{\vspace*{0.65cm}}{}{}{%
    \PackageError{rlec}{Could not remove the unused title metadata space}{}
}

\newcommand{\rlecpatchtitle}[3]{%
    \patchcmd{#1}{#2}{#3}{}{%
        \PackageError{rlec}{Could not compact Seed title spacing}{}
    }%
}
\makeatletter
\rlecpatchtitle{\@toptitlebar}{\vskip 6mm}{\vskip 3.5mm}
\rlecpatchtitle{\@bottomtitlebar}{\vskip 6mm}{\vskip 3.5mm}
\makeatother
\rlecpatchtitle{\mymaketitle}{\tcbset{top=6mm}}{\tcbset{top=3.5mm}}
\rlecpatchtitle{\mymaketitle}{\tcbset{bottom=6mm}}{\tcbset{bottom=3.5mm}}
\rlecpatchtitle{\mymaketitle}{\vskip 6mm}{\vskip 3mm}
\rlecpatchtitle{\mymaketitle}{\vskip 6mm}{\vskip 3mm}
\rlecpatchtitle{\maketitle}{\vskip 8mm}{\vskip 3mm}

\newtcolorbox{promptcard}[1]{
    enhanced,
    breakable,
    colback=gray!4,
    colframe=black!55,
    boxrule=0.6pt,
    arc=1mm,
    left=6pt,
    right=6pt,
    top=5pt,
    bottom=5pt,
    title={#1},
    fonttitle=\bfseries,
    fontupper=\small\ttfamily
}

\title{Reinforcement Learning Done Right for Explicit Confidence Targets}
\author[1,*]{Zefeng He}
\author[1]{Yichen He}
\author[1,\dagger]{Yuan Lin}
\author[2,\dagger]{Yang Yang}
\author[3]{Yu Cheng}

\affiliation[1]{ByteDance Seed}
\affiliation[2]{Shanghai Jiao Tong University}
\affiliation[3]{NTU}

\contribution[*]{Work done during an internship at Bytedance Seed}
\contribution[\dagger]{Corresponding authors}

\abstract{Large language models (LLMs) can produce plausible but incorrect answers when they are uncertain, which can mislead users and undermine model reliability. Training models to abstain with reinforcement learning is a natural approach, but most existing methods use a fixed error penalty and therefore cannot adapt their behavior to different confidence requirements at inference time. We study \emph{explicit confidence targets}, where a user specifies a target \(p\) in the prompt and the model directly chooses whether to answer or abstain, with the goal of achieving an accuracy of at least \(p\) among answered questions. However, learning to respond to different confidence targets with a single model has proven difficult, and prior attempts have failed to distinguish different targets. We show that this failure comes from two problems: the reward signal that distinguishes different values of $p$ is often lost during optimization, while the original error penalty grows without bound as $p$ approaches one. We propose \textbf{RLEC} (\textbf{R}einforcement \textbf{L}earning with \textbf{E}xplicit \textbf{C}onfidence Targets), which preserves the learning signal needed to distinguish confidence targets, introduces a stable reward family that keeps the desired answer condition unchanged, and uses reward scheduling to balance effective learning and training stability. Together, these designs enable the first successful training of a single LLM for explicit confidence target control. Extensive experiments across diverse benchmarks and model scales show that RLEC largely preserves general capabilities and achieves competitive calibration performance. RLEC also supports fine-grained target control without an external decision system or threshold calibration after training. By making explicit confidence targets part of the trained policy, RLEC opens a new and more direct path toward behavioral calibration.
}
\date{October 26, 2026}
\checkdata[Code]{\url{https://github.com/ByteDance-Seed/RLEC}}

\begin{document}
\maketitle

\section{Introduction}
\label{sec:introduction}

\begingroup
\setlength{\intextsep}{4pt}
\begin{wrapfigure}[12]{r}{0.49\textwidth}
    \centering
    \includegraphics[width=\linewidth]{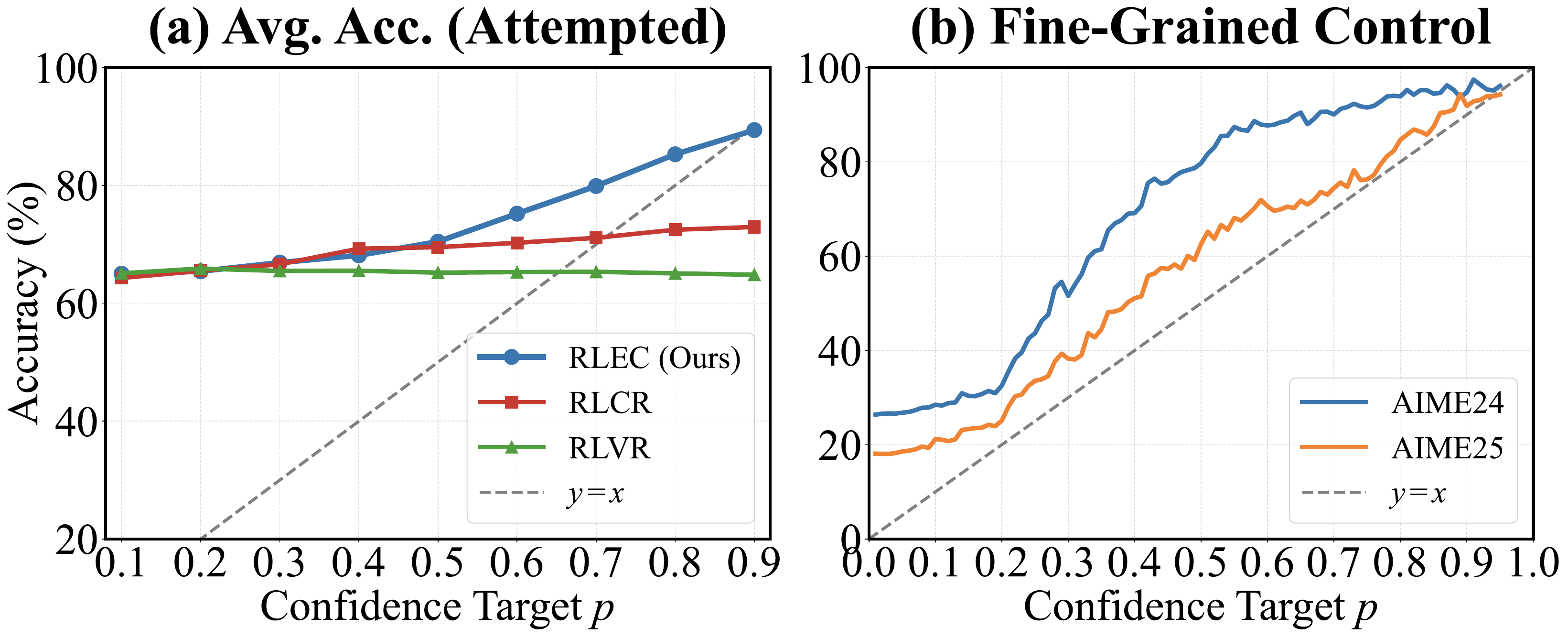}
    \par
    \microtypesetup{protrusion=false}
    \captionsetup{font=small,skip=6pt,
        justification=justified,singlelinecheck=false,margin=0pt,indention=0pt}
    \caption{\textbf{Explicit Confidence Target control.} (a)~RLEC's average accuracy across eight benchmarks increases with \(p\) specified in the prompt. (b)~RLEC's fine-grained control on AIME with \(p\).}
    \label{fig:intro_results}
\end{wrapfigure}

Large language models (LLMs)~\cite{achiam2023gpt,comanici2025gemini,team2026kimi} have shown strong abilities in knowledge-intensive and reasoning tasks, but they can still generate plausible yet incorrect responses, commonly known as hallucinations~\cite{ji2023survey,huang2025survey}. Such errors reduce the reliability of LLMs and may mislead users, posing particular risks when incorrect answers are costly. Rather than guessing when uncertain, a reliable model should know when not to answer~\cite{kamath2020selective,kadavath2022language,kalai2025language}.

\leavevmode\citet{kalai2025language} argue that common training objectives reward guessing over acknowledging uncertainty. A natural solution is to train abstention with reinforcement learning~\cite{wei2025truthrl,mohamadi2025honesty,jha2026rewarding,pan2026tiar}, rewarding correct answers, penalizing incorrect answers, and assigning no reward to abstentions. For example, TruthRL~\cite{wei2025truthrl} uses such a reward with a fixed error penalty. However, a fixed penalty represents only one risk preference, while different applications may require different levels of reliability. To represent these requirements, we study an \emph{explicit confidence target}: the user specifies a target \(p\) in the prompt, and the model directly chooses whether to answer or abstain. The desired behavior is
\begin{equation}
    \mathrm{Acc}(p)
    =
    \Pr(\mathrm{correct}\mid\mathrm{answer},p)
    \geq p,
    \label{eq:intro_target}
\end{equation}
and a single model should respond to different values of \(p\) without retraining or an external decision system. However, learning such behavior has proved difficult. ~\citet{wu2025mitigating} propose Explicit Risk Thresholding (ERT), which trains on randomly sampled values of \(p\) using GRPO~\cite{deepseekmath,deepseekr1}, assigning rewards of \(1\), \(-p/(1-p)\), and \(0\) to correct answers, incorrect answers, and abstentions, respectively. However, they report that the model's hallucination rates remain insensitive to the input target. They attribute this failure to the noisy optimization problem created by training across many thresholds. Other work~\cite{mohamadi2025honesty} further argues that limited model capacity may force a policy to specialize in one risk preference. These findings leave it unclear whether the difficulty is inherent to a single model or caused by the training method.
\par\endgroup

We show that this failure does not mean that the task is beyond the capacity of a single model. Instead, it can be explained by two problems in the training setup. First, ERT, like many recent RL methods for LLMs, uses GRPO. We observe that most of its rollout groups contain only one or two distinct reward values. Groups with a single reward value yield zero advantages, while in groups with two values, standard deviation normalization removes the effect of the reward scale induced by \(p\) on the advantages. The resulting advantages depend only on the fraction of the two values, rather than their magnitudes. As a result, reward magnitudes provide little direct signal for distinguishing confidence targets. Second, the original error reward grows without bound as \(p\) approaches one. Once the reward magnitudes directly affect the advantages, samples with large \(p\) can dominate the gradient and destabilize training. Based on these findings, we introduce \textbf{RLEC} (\textbf{R}einforcement \textbf{L}earning with \textbf{E}xplicit \textbf{C}onfidence Targets). RLEC removes division by the group standard deviation while retaining mean centering, thereby preserving the effect of the reward scale induced by \(p\) on the advantages. It further introduces a reward family controlled by a stability coefficient \(b\), which controls reward magnitudes at large \(p\) without changing the desired decision rule. Finally, we observe that fixed values of \(b\) tend to trade off the strength of the learning signal against training stability. We therefore introduce a schedule for \(b\) that combines a strong signal for learning different confidence targets with stable training.

We evaluate RLEC using Qwen3-4B-Base and Qwen3-30B-A3B-Base~\cite{yang2025qwen3} across eight benchmarks covering mathematical reasoning, knowledge and hallucination, and scientific reasoning. Our results show that a single model can successfully learn different behaviors for different explicit confidence targets, as shown in Figure~\ref{fig:intro_results}(a). We also compare RLEC with a standard RLVR~\cite{deepseekr1} baseline, showing that RLEC largely preserves the baseline's general capabilities at both model scales. For additional context, we further compare RLEC with RLCR~\cite{damani2026beyond}, which focuses on confidence reporting rather than direct abstention control. RLEC achieves calibration performance comparable to RLCR at 4B and obtains a lower overconfidence rate and a higher overall score at 30B. RLEC also provides several practical advantages. Methods based on verbalized confidence~\cite{tian2023just,damani2026beyond} often require an external rule to convert their confidence estimates into answer or abstention decisions, while other selective prediction methods~\cite{wang2026safer,kamath2020selective} further rely on separate calibrators or held-out calibration sets. In contrast, RLEC receives \(p\) in the prompt and directly produces the final answer or abstention without external processing. It also supports fine-grained control over answer or abstention, with smooth changes across densely spaced values of \(p\), as illustrated in Figure~\ref{fig:intro_results}(b).

Our main contributions are as follows:
\begin{itemize}[leftmargin=*]
    \item We identify how the optimization algorithm and reward design used in prior work weaken the confidence target signal, providing a concrete explanation for the previously reported failure.
    \item We propose RLEC, which preserves the confidence target signal, introduces a stable reward family that keeps the desired answer condition unchanged, and uses reward scheduling to  balance effective learning and training stability. To our knowledge, RLEC thereby enables the first successful training of a single LLM for explicit confidence target control.
    \item Extensive experiments across diverse benchmarks and model scales show that RLEC largely preserves general capabilities and achieves competitive calibration performance. It also requires no external decision system or calibration after training and supports fine-grained target control.
\end{itemize}

\section{Background}
\label{sec:background}

\subsection{Fixed-Penalty Abstention}
\label{sec:fixed_penalty}

To mitigate hallucinations \cite{ji2023survey,huang2025survey}, a reliable model should know when to answer and when to abstain. Given a question \(x\), the model chooses an action \(a\in\{\mathrm{ANS},\mathrm{ABS}\}\), where \(\mathrm{ANS}\) denotes answering and \(\mathrm{ABS}\) denotes abstaining. We primarily evaluate on verifiable question answering tasks following prior work~\cite{wei2025truthrl,damani2026beyond}. This setup yields three outcomes: a correct answer, an incorrect answer, or an abstention. We treat an incorrect answer as a hallucination. We define accuracy among answered questions as
\begin{equation}
    \mathrm{Acc}
    =
    \Pr(\mathrm{correct}\mid a=\mathrm{ANS}).
    \label{eq:accuracy}
\end{equation}

Recent work~\cite{wei2025truthrl,mohamadi2025honesty,jha2026rewarding} has used reinforcement learning with verifiable rewards (RLVR)~\cite{deepseekmath,deepseekr1} to directly train LLMs to decide whether to answer or abstain. A common formulation assigns a positive reward to a correct answer, a neutral reward to abstention, and a fixed negative reward to an incorrect answer:
\begin{equation}
R_{\lambda}(a,y)=
\begin{cases}
1, & a=\mathrm{ANS}\ \text{and } y \text{ is correct},\\
0, & a=\mathrm{ABS},\\
-\lambda, & a=\mathrm{ANS}\ \text{and } y \text{ is incorrect},
\end{cases}
\qquad \lambda \geq 0.
\label{eq:fixed_ternary_reward}
\end{equation}
TruthRL \cite{wei2025truthrl} optimizes this reward with \(\lambda=1\) using Group Relative Policy Optimization (GRPO)~\cite{deepseekmath}, whereas Reinforced Hesitation \cite{mohamadi2025honesty} trains separate models for different values of \(\lambda\).
Let \(q(x)\) be the true conditional probability that the model's intended answer to \(x\) is correct. Under Equation~\ref{eq:fixed_ternary_reward}, the expected reward of answering is
\begin{equation}
\begin{aligned}
\mathbb{E}\!\left[R_{\lambda}\mid x,a=\mathrm{ANS}\right]
&=q(x)-\lambda\bigl(1-q(x)\bigr)\\
&=(1+\lambda)q(x)-\lambda.
\end{aligned}
\label{eq:fixed_reward_expectation}
\end{equation}
Since abstention yields zero reward, the policy answers when
\begin{equation}
    q(x)>\frac{\lambda}{1+\lambda},
    \label{eq:fixed_confidence_threshold}
\end{equation}
and abstains when the inequality is reversed. Equivalently, \(\lambda\) determines the fixed confidence threshold
\begin{equation}
    p_{\lambda}=\frac{\lambda}{1+\lambda}.
\end{equation}

Under the idealized assumptions that \(q(x)\) gives the true probability of correctness and that the policy exactly follows Equation~\ref{eq:fixed_confidence_threshold}, its selective accuracy satisfies
\begin{equation}
\begin{aligned}
\Pr(\mathrm{correct}\mid a=\mathrm{ANS})
&=
\mathbb{E}\!\left[q(x)\mid a=\mathrm{ANS}\right]\\
&=
\mathbb{E}\!\left[q(x)\mid q(x)>p_{\lambda}\right]
\geq p_{\lambda}.
\end{aligned}
\label{eq:fixed_selective_accuracy}
\end{equation}
This characterizes idealized behavior, not an empirical guarantee, since the learned policy may not optimize or generalize perfectly and its uncertainty may not match the true probability of correctness.

Training with a fixed penalty provides an interpretable trade-off between accuracy and coverage, but each \(\lambda\) represents only one operating point. Adapting to different error costs therefore generally requires retraining the model~\cite{mohamadi2025honesty,jha2026rewarding} or selecting a different model.

\subsection{Explicit Confidence Targets}
\label{sec:explicit_confidence_targets}

Explicit confidence targets~\cite{kalai2025language} extend fixed-penalty abstention by allowing users to specify the required accuracy in the prompt. Let \(p\in[0,1)\) denote this target. Setting
\begin{equation}
    \lambda(p)=\frac{p}{1-p}
    \label{eq:penalty_target_relation}
\end{equation}
in Equation~\ref{eq:fixed_ternary_reward} gives
\begin{equation}
R_p=
\begin{cases}
1, & \text{correct answer},\\
-\dfrac{p}{1-p}, & \text{incorrect answer},\\
0, & \text{abstention}.
\end{cases}
\label{eq:explicit_target_reward}
\end{equation}
Suppose the probability of answering correctly is \(q(x)\). The expected reward for answering is
\begin{equation}
\begin{aligned}
\mathbb{E}\!\left[R_p\mid x,\mathrm{ANS}\right]
&=q(x)-\frac{p}{1-p}\bigl(1-q(x)\bigr)\\
&=\frac{q(x)-p}{1-p}.
\end{aligned}
\label{eq:explicit_reward_expectation}
\end{equation}
Abstention receives zero reward. A policy that maximizes expected reward therefore answers when \(q(x)>p\), abstains when \(q(x)<p\), and is indifferent when \(q(x)=p\). Thus, \(p\) directly controls the decision between answering and abstaining. When \(p=0\), answering is at least as good as abstaining for every question and is strictly better whenever \(q(x)>0\). We use this endpoint to represent the setting in which the user requires the model to answer, with ties resolved in favor of answering.

Our goal is to train a single policy \(\pi_{\theta}(\cdot\mid x,p)\) that follows arbitrary confidence targets without retraining. For every \(p\) at which the model answers with nonzero probability, the desired behavior is
\begin{equation}
    \mathrm{Acc}(p)
    :=
    \Pr(\mathrm{correct}\mid \mathrm{ANS},p)
    \geq p.
    \label{eq:explicit_calibration_objective}
\end{equation}

The model should answer when doing so yields positive expected reward, rather than defaulting to abstention on every question. Importantly, \(p\) is a target specified by the user in the prompt, and the model directly chooses whether to answer or abstain. Inference therefore requires no external system to compare a reported confidence score~\cite{damani2026beyond} with a threshold.

\citet{wu2025mitigating} propose \emph{Explicit Risk Thresholding} (ERT) to study this approach by sampling \(p\) during training, including it in the prompt, and using the reward in Equation~\ref{eq:explicit_target_reward}. However, their hallucination rates remain largely unchanged across different \(p\), showing that the model does not adapt its behavior to the target. They attribute this failure to a noisy optimization problem caused by sampling targets across the full range, making it difficult to learn a consistent strategy for different \(p\). ~\citet{mohamadi2025honesty} also argue that a single model cannot support all risk preferences because limited model capacity prevents it from optimizing every error penalty simultaneously. Together, these findings illustrate the difficulty of training one model to follow arbitrary confidence targets.

\section{Method}
\label{sec:method}

\subsection{Signal Preservation}
\label{sec:signal_preservation}

We first identify a key problem in prior work: most of the direct reward signal distinguishing different confidence targets is lost during training. ERT uses GRPO to optimize the reward in Equation~\ref{eq:explicit_target_reward}. For each prompt, GRPO samples \(G\) responses and estimates their advantages from relative rewards. Let \(R_i\) denote the reward of response \(i\). The corresponding advantage is
\begin{equation}
A_i=\frac{R_i-\overline{R}}{\sigma_R},
\qquad
\overline{R}=\frac{1}{G}\sum_{j=1}^{G}R_j,
\qquad
\sigma_R=\sqrt{\frac{1}{G}\sum_{j=1}^{G}(R_j-\overline{R})^2}.
\label{eq:grpo_advantage}
\end{equation}

However, this standardization can remove the effect of the reward scale induced by \(p\) on the advantages. At a fixed \(p\), consider a rollout group with two distinct reward values \(r_{\mathrm{H}}>r_{\mathrm{L}}\). Let \(\rho\in(0,1)\) denote the fraction of responses receiving \(r_{\mathrm{H}}\). The group mean and standard deviation are
\begin{equation}
\overline{R}=\rho r_{\mathrm{H}}+(1-\rho)r_{\mathrm{L}},
\qquad
\sigma_R=(r_{\mathrm{H}}-r_{\mathrm{L}})\sqrt{\rho(1-\rho)}.
\label{eq:two_value_statistics}
\end{equation}
Substituting Equation~\ref{eq:two_value_statistics} into Equation~\ref{eq:grpo_advantage} gives
\begin{equation}
A_{\mathrm{H}}=\sqrt{\frac{1-\rho}{\rho}},
\qquad
A_{\mathrm{L}}=-\sqrt{\frac{\rho}{1-\rho}}.
\label{eq:two_value_advantage}
\end{equation}

The resulting advantages are independent of \(p\), so the training signal carried by the reward scale of \(p\) disappears. For example, in a group with two distinct reward values, a higher \(p\) imposes a larger reward penalty on an incorrect answer, but does not make its standardized advantage more negative.
Also, for a group with one reward value, all advantages in Equation~\ref{eq:grpo_advantage} are computed as zero in our implementation, so the group provides no reward signal. Only groups containing all three outcomes can retain the learning signal from changes in the reward scale induced by \(p\).

\begin{wrapfigure}{r}{0.48\textwidth}
    \centering
    \includegraphics[width=\linewidth]{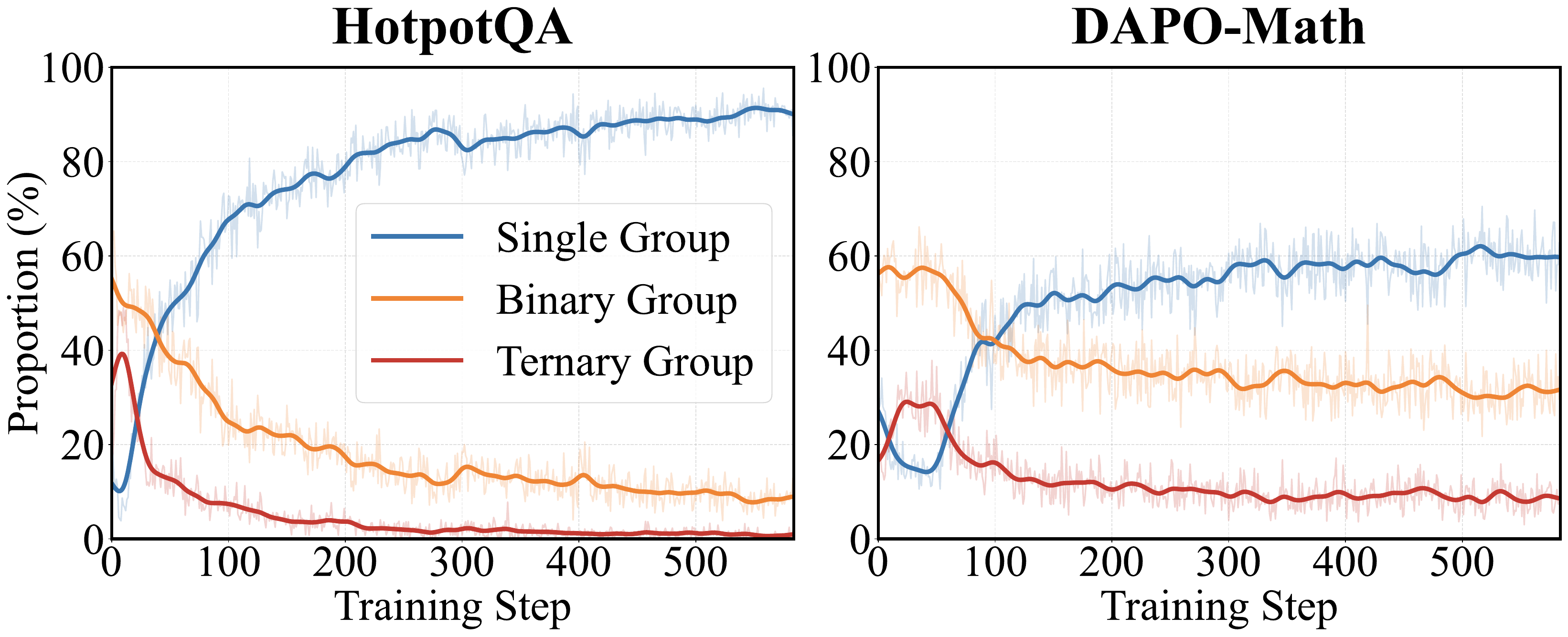}
    \caption{\textbf{Fractions of rollout groups.}}
    \label{fig:group_composition}
\end{wrapfigure}

Figure~\ref{fig:group_composition} shows how the three group types evolve during training. Groups with three reward values are rare and become less frequent, while groups with one value eventually dominate and most others contain two. 
Thus, only a small fraction of groups retains a direct signal from reward magnitudes that vary with \(p\), explaining why the policy remains insensitive to the confidence target.

To preserve the signal, we remove the group standard deviation while retaining mean centering:
\begin{equation}
\widetilde{A}_i=R_i-\overline{R}.
\label{eq:mean_only_advantage}
\end{equation}
Without standard deviation normalization, a reward gap that varies with \(p\) produces advantages that vary accordingly, preserving the direct reward signal across confidence targets. This also exposes the large penalty in Equation~\ref{eq:explicit_target_reward}, which we address next.

\subsection{Stable Rewards}
\label{sec:stable_rewards}

Removing the group standard deviation preserves the effect of reward magnitudes on the advantages, but also makes policy updates sensitive to those magnitudes. In Equation~\ref{eq:explicit_target_reward}, the rewards for a correct answer and abstention remain fixed at \(1\) and \(0\), while the reward for an incorrect answer is \(-p/(1-p)\), which tends to \(-\infty\) as \(p\) approaches one. Groups containing both incorrect answers and other outcomes can therefore produce large advantages. This creates two related problems.

\paragraph{Gradient dominance.} Because the policy shares parameters across confidence targets, such groups at large \(p\) can dominate the overall gradient, reducing the influence of lower targets.

\paragraph{Training instability.} The number and composition of such groups vary across batches. At large \(p\), their centered advantages can cause sharp fluctuations in gradient norms and destabilize training.


Multiplying all rewards by a single constant scales the centered advantages equally across confidence targets, leaving their relative influence unchanged. We instead use a positive factor that decreases with \(p\) to reduce the penalty at large \(p\) without changing the theoretical decision to answer or abstain.

A simple way to bound the error penalty is to multiply all rewards by \(1-p\):
\begin{equation}
(1-p)R_p=
\begin{cases}
1-p, & \text{correct answer},\\
-p, & \text{incorrect answer},\\
0, & \text{abstention}.
\end{cases}
\label{eq:unit_gap_reward}
\end{equation}
The magnitude of the error penalty is now bounded by one. This scaling also changes the reward for a correct answer to \(1-p\), while abstention remains at zero.

To control how strongly rewards at large \(p\) are reduced, we introduce a family of rewards:
\begin{equation}
R_p^{(b)}=(1-p)^bR_p=
\begin{cases}
(1-p)^b, & \text{correct answer},\\
-p(1-p)^{b-1}, & \text{incorrect answer},\\
0, & \text{abstention},
\end{cases}
\qquad b\geq0.
\label{eq:stable_reward}
\end{equation}
Here, \(b\) is a stability coefficient. For a fixed \(p\in(0,1)\), increasing \(b\) reduces the magnitude of the error penalty and the reward for a correct answer. In particular, \(b=0\) recovers Equation~\ref{eq:explicit_target_reward}, while \(b=1\) recovers Equation~\ref{eq:unit_gap_reward}. Since the same scaling applies to all responses in a rollout group, it also scales their centered advantages:
\begin{equation}
\widetilde{A}_i^{(b)}
=(1-p)^b\widetilde{A}_i^{(0)}.
\label{eq:stable_advantage_scaling}
\end{equation}
Thus, larger values of \(b\) reduce the relative influence of groups with large \(p\).

Importantly, scaling all rewards at a given \(p\) by the same positive factor does not change the theoretical decision to answer or abstain. Suppose the model answers correctly with probability \(q(x)\). Its expected reward for answering is
\begin{equation}
\begin{aligned}
\mathbb{E}\!\left[R_p^{(b)}\mid x,\mathrm{ANS}\right]
&=q(x)(1-p)^b-\bigl(1-q(x)\bigr)p(1-p)^{b-1}\\
&=(1-p)^{b-1}\bigl(q(x)-p\bigr).
\end{aligned}
\label{eq:stable_reward_expectation}
\end{equation}
Since \((1-p)^{b-1}>0\) for \(p<1\), answering has positive expected reward exactly when \(q(x)>p\). Thus, every value of \(b\) preserves the same theoretical answer condition.

\subsection{Reward Scheduling}
\label{sec:reward_scheduling}

The stability coefficient \(b\) balances the strength of the learning signal against training stability. For any fixed \(p\in(0,1)\), the centered advantages in Equation~\ref{eq:stable_advantage_scaling} become smaller in magnitude as \(b\) increases. This reduction is strongest at large \(p\), where the original error penalty is largest. It can improve stability, but it also weakens the signal for learning when to abstain by reducing the error penalty relative to abstention. Our ablations suggest that keeping \(b\) fixed at a relatively large value, such as \(b=1\), can make it harder for the model to learn when to abstain and to vary its behavior across confidence targets. Details are shown in Section~\ref{sec:ablation}.

By contrast, a smaller \(b\), such as \(b=0\), preserves stronger centered advantages at large \(p\), helping the model learn when to abstain and distinguish confidence targets more effectively early in training. However, keeping \(b\) small throughout training can lead to gradient dominance and training instability, as discussed in Section~\ref{sec:stable_rewards}. These effects may harm final performance and general capabilities.

Guided by this trade-off, we linearly increase \(b\) from \(0\) to \(1\) over \(T\) training steps. At step \(t\), we set
\begin{equation}
b_t=\frac{t}{T},
\qquad 0\leq t\leq T.
\label{eq:b_schedule}
\end{equation}
This schedule preserves a stronger learning signal early while improving training stability later.
\section{Experiment}
\label{sec:experiments}

In this section, we conduct experiments to answer the following research questions:
\begin{itemize}[leftmargin=*, itemsep=0pt, topsep=2pt]
\item \textbf{(RQ1)} Can RLEC successfully distinguish different explicit confidence targets?
\item \textbf{(RQ2)} How does RLEC compare with prior methods in calibration and general capabilities?
\item \textbf{(RQ3)} What enables RLEC to distinguish targets while preserving general capabilities?
\item \textbf{(RQ4)} What other properties and practical advantages does RLEC offer?
\end{itemize}

\subsection{Experimental Setup}
\label{sec:experimental_setup}

\paragraph{Implementation Details.} Our main experiments and ablation studies are based on Qwen3-4B-Base~\cite{yang2025qwen3}, and we further conduct experiments on Qwen3-30B-A3B-Base to evaluate scalability. Our training data combine DAPO-Math~\cite{yu2026dapo} and a modified version of HotpotQA~\cite{yang2018hotpotqa,damani2026beyond}. We randomly sample 15,000 examples from each dataset, resulting in 30,000 training examples, and train for five epochs. We set \(p=0\) with probability \(10\%\). Otherwise, we uniformly sample \(p\) from the interval \([0,0.9]\) and round it to two decimal places. During evaluation, we test \(p\in\{0,0.1,\ldots,0.9\}\) and report results averaged over multiple sampled generations. Additional details are provided in Appendix~\ref{app:implementation_details}.

\paragraph{Benchmarks.} Our evaluation covers eight benchmarks from three domains and different difficulty levels, including \textbf{(1) mathematical reasoning:} GSM8K~\cite{cobbe2021training}, MATH-500~\cite{hendrycks2021measuring,lightman2024let}, and AIME 2024; \textbf{(2) knowledge and hallucination:} HotpotQA test set~\cite{yang2018hotpotqa,damani2026beyond}, AA-Omniscience~\cite{jackson2025aa}, and SimpleQA-Verified~\cite{wei2024measuring,haas2025simpleqa}; and \textbf{(3) scientific reasoning:} MMLU-Pro~\cite{wang2024mmlu} and GPQA~\cite{rein2023gpqa}. Details are provided in Appendix~\ref{app:benchmark_details}.

\paragraph{Baselines.} We compare RLEC with three types of methods: \textbf{(1) RLVR:} standard RLVR with a binary reward; \textbf{(2) abstention methods:} methods that use a fixed error penalty or fail to learn behavior that varies with \(p\), including TruthRL~\cite{wei2025truthrl} and ERT~\cite{wu2025mitigating}, compared in RQ1; and \textbf{(3) confidence reporting methods:} RLCR~\cite{damani2026beyond}, included in RQ2 as an additional reference for calibration performance under a different formulation.

\subsection{Adaptation to Confidence Targets}
\label{sec:target_adaptation}

To answer \textbf{RQ1}, we study whether models can change their behavior according to \(p\). We first compare the training behavior of RLEC and ERT, and then compare the final accuracy and abstention rates of RLEC, RLVR, ERT, and TruthRL across all evaluation benchmarks.

\begin{figure}[t]
    \centering
    \includegraphics[width=\textwidth]{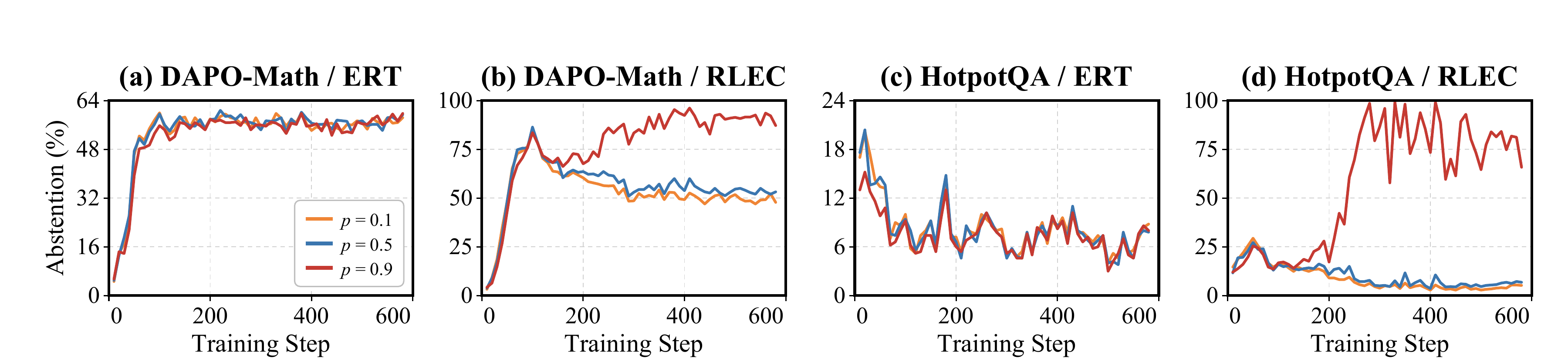}
    \caption{\textbf{Abstention rates on held-out validation sets during training.} Each curve evaluates the same training checkpoints with a fixed confidence target \(p\). ERT produces nearly overlapping curves for different targets, while RLEC learns higher abstention rates for higher values of \(p\).}
    \label{fig:training_curves1}
\end{figure}

\paragraph{Training Curves.} Figure~\ref{fig:training_curves1} shows the abstention rate on held-out validation sets during training, evaluated with \(p\in\{0.1,0.5,0.9\}\). On both HotpotQA and DAPO-Math, our reproduction of ERT shows the same failure reported in the original study~\cite{wu2025mitigating}: its curves for different values of \(p\) almost completely overlap. In contrast, RLEC produces a clear dependence on \(p\), with higher confidence targets generally leading to higher abstention rates. These results show that RLEC successfully learns different behaviors for different confidence targets.

\begin{figure}[t]
    \centering
    \includegraphics[width=\textwidth]{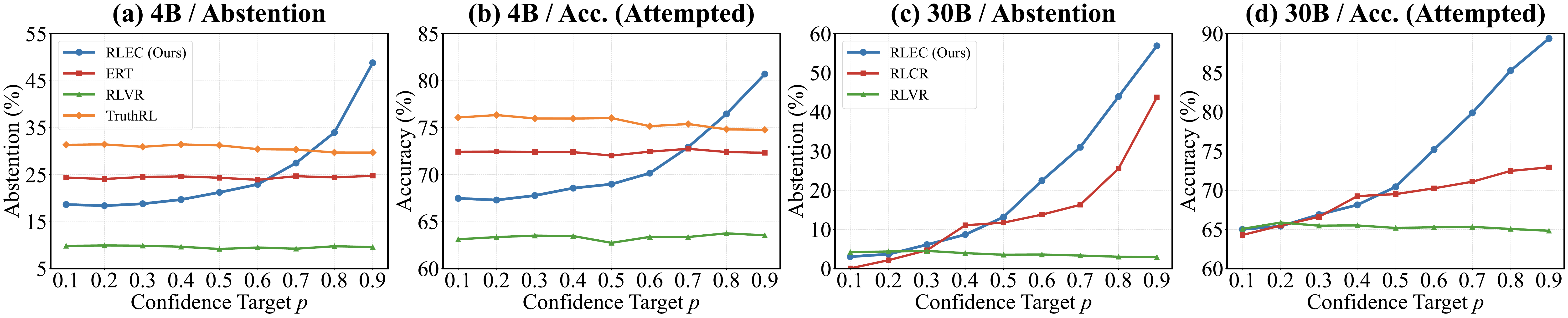}
    \caption{\textbf{Adaptation to explicit confidence targets.} (a,b) Direct decision methods with Qwen3-4B-Base. (c,d) RLEC, RLCR, and RLVR with Qwen3-30B-A3B-Base.}
    \label{fig:target_control}
\end{figure}

\paragraph{Test Results.} Figure~\ref{fig:target_control}(a,b) reports average abstention rates and answer accuracy across all benchmarks using Qwen3-4B-Base. Among methods that directly decide whether to answer, only RLEC changes clearly with \(p\): as \(p\) increases from \(0.1\) to \(0.9\), its abstention rate rises from approximately \(19\%\) to \(49\%\), while its accuracy among answered questions rises from approximately \(68\%\) to \(81\%\). RLVR, ERT, and TruthRL remain nearly unchanged. Thus, RLEC successfully learns to adapt its decision to answer or abstain across confidence targets, answering RQ1.

\subsection{Comparison with Calibration Methods}
\label{sec:calibration_comparison}

As shown in Section~\ref{sec:target_adaptation}, existing methods that directly decide whether to answer or abstain show little sensitivity to the confidence target. To compare calibration performance and answer \textbf{RQ2}, we therefore compare RLEC with RLCR~\cite{damani2026beyond}, a different type of method that reports a numerical confidence \(c\) together with its answer. Since RLCR cannot directly follow a user provided target \(p\), we apply an external rule that retains the answer when \(c\geq p\) and replaces it with an abstention otherwise, whereas RLEC conditions its answer or abstention decision directly on \(p\).

\paragraph{Qualitative Analysis.} Figure~\ref{fig:target_control}(c,d) compares RLEC, RLCR, and RLVR across confidence targets using Qwen3-30B-A3B-Base. RLVR changes little with \(p\), whereas RLEC and RLCR increase their abstention rates as \(p\) increases. RLEC responds more strongly, especially at high \(p\), and its accuracy among answered questions follows the target boundary more closely. These results suggest that directly optimizing explicit confidence targets enables clearer adaptation to the requested target.

\paragraph{Quantitative Analysis.} We use three metrics to evaluate answer ability, target violations, and utility. \emph{Forced Accuracy} measures answer accuracy when abstention is disabled. We set \(p=0\) for RLEC and retain all RLCR answers, testing whether calibration training harms answer ability.

\emph{Average Overconfidence} (AOC) measures the average magnitude of target violations:
\begin{equation}
    \mathrm{AOC}
    =
    \frac{1}{|\mathcal{P}|}
    \sum_{p\in\mathcal{P}}
    \max\left\{0,\;p-\mathrm{Acc}(p)\right\},
    \label{eq:average_overconfidence}
\end{equation}
where lower values indicate better compliance with the requested accuracy. Since AOC can be reduced by abstaining, it does not capture utility. We therefore also report
\begin{equation}
    \mathrm{Score}
    =
    \frac{1}{N|\mathcal{P}|}
    \sum_{p\in\mathcal{P}}
    \sum_{i=1}^{N}
    z_{i,p}\left(y_{i,p}-p\right),
    \label{eq:average_selective_score}
\end{equation}
where \(z_{i,p}\) indicates whether the model answers and \(y_{i,p}\) indicates whether the answer is correct. A correct answer contributes \(1-p\), an incorrect answer contributes \(-p\), and an abstention contributes zero. A higher Score indicates greater utility by accounting for both correctness and coverage.

For RLEC, each Score contribution is exactly its reward with \(b=1\). For RLCR, twice the corresponding score integrated over all confidence targets equals its reward, as proved in Appendix~\ref{app:score_equivalence}.

\begin{table}[t]
    \centering
    \caption{Forced answer accuracy, average overconfidence (AOC), and average score across model scales. The best Micro and Macro results within each model scale are shown in bold.}
    \label{tab:calibration_results}
    \resizebox{\textwidth}{!}{%
    \begin{tabular}{@{}l|llcccccccccc@{}}
        \toprule
        & & & \multicolumn{3}{c}{Mathematical Reasoning} & \multicolumn{3}{c}{Knowledge \& Hallucination} & \multicolumn{2}{c}{Scientific Reasoning} & \multicolumn{2}{c}{Avg.} \\
        \cmidrule(lr){4-6}\cmidrule(lr){7-9}\cmidrule(lr){10-11}\cmidrule(lr){12-13}
        Model & Metric & Method & GSM8K & MATH & AIME & Hotpot & OmniSci & SimpleQA & MMLU-Pro & GPQA & Micro & Macro \\
        \midrule
        \multirow{9}{*}{\shortstack{Qwen3-4B-\\Base}} & \multirow{3}{*}{Acc. $\uparrow$} & RLVR & 93.33 & 69.70 & 20.83 & 69.50 & 13.75 & 5.20 & 68.47 & 35.91 & \textbf{58.18} & \textbf{47.09} \\
         &  & RLCR & 93.10 & 73.90 & 16.25 & 66.30 & 15.58 & 4.75 & 66.94 & 34.00 & 57.35 & 46.35 \\
         &  & \cellcolor{gray!12}RLEC & \cellcolor{gray!12}92.15 & \cellcolor{gray!12}70.50 & \cellcolor{gray!12}19.79 & \cellcolor{gray!12}67.60 & \cellcolor{gray!12}14.17 & \cellcolor{gray!12}4.75 & \cellcolor{gray!12}67.37 & \cellcolor{gray!12}36.24 & \cellcolor{gray!12}57.40 & \cellcolor{gray!12}46.57 \\
        \cmidrule(lr){2-13}
         & \multirow{3}{*}{AOC $\downarrow$} & RLVR & 0.000 & 0.027 & 0.258 & 0.033 & 0.292 & 0.389 & 0.032 & 0.175 & 0.049 & 0.151 \\
         &  & RLCR & 0.090 & 0.090 & 0.090 & 0.098 & 0.229 & 0.396 & 0.090 & 0.169 & 0.090 & 0.156 \\
         &  & \cellcolor{gray!12}RLEC & \cellcolor{gray!12}0.000 & \cellcolor{gray!12}0.000 & \cellcolor{gray!12}0.090 & \cellcolor{gray!12}0.010 & \cellcolor{gray!12}0.259 & \cellcolor{gray!12}0.381 & \cellcolor{gray!12}0.015 & \cellcolor{gray!12}0.157 & \cellcolor{gray!12}\textbf{0.013} & \cellcolor{gray!12}\textbf{0.114} \\
        \cmidrule(lr){2-13}
         & \multirow{3}{*}{Score $\uparrow$} & RLVR & 0.480 & 0.261 & -0.162 & 0.237 & -0.173 & -0.213 & 0.237 & -0.090 & 0.167 & 0.072 \\
         &  & RLCR & 0.482 & 0.342 & 0.053 & 0.255 & -0.004 & -0.094 & 0.285 & 0.069 & \textbf{0.235} & \textbf{0.173} \\
         &  & \cellcolor{gray!12}RLEC & \cellcolor{gray!12}0.477 & \cellcolor{gray!12}0.333 & \cellcolor{gray!12}0.056 & \cellcolor{gray!12}0.264 & \cellcolor{gray!12}-0.048 & \cellcolor{gray!12}-0.107 & \cellcolor{gray!12}0.262 & \cellcolor{gray!12}-0.016 & \cellcolor{gray!12}0.214 & \cellcolor{gray!12}0.153 \\
        \midrule
        \multirow{9}{*}{\shortstack{Qwen3-30B-\\A3B-Base}} & \multirow{3}{*}{Acc. $\uparrow$} & RLVR & 95.07 & 77.10 & 26.88 & 68.90 & 18.17 & 9.40 & 76.25 & 46.53 & 63.97 & 52.29 \\
         &  & RLCR & 94.73 & 77.70 & 26.25 & 66.20 & 19.08 & 8.90 & 77.22 & 46.76 & \textbf{64.24} & 52.10 \\
         &  & \cellcolor{gray!12}RLEC & \cellcolor{gray!12}94.77 & \cellcolor{gray!12}78.60 & \cellcolor{gray!12}27.29 & \cellcolor{gray!12}66.60 & \cellcolor{gray!12}20.00 & \cellcolor{gray!12}8.30 & \cellcolor{gray!12}76.73 & \cellcolor{gray!12}47.09 & \cellcolor{gray!12}64.14 & \cellcolor{gray!12}\textbf{52.42} \\
        \cmidrule(lr){2-13}
         & \multirow{3}{*}{AOC $\downarrow$} & RLVR & 0.000 & 0.014 & 0.221 & 0.038 & 0.265 & 0.372 & 0.020 & 0.108 & 0.045 & 0.130 \\
         &  & RLCR & 0.000 & 0.003 & 0.000 & 0.038 & 0.270 & 0.367 & 0.000 & 0.057 & 0.025 & 0.092 \\
         &  & \cellcolor{gray!12}RLEC & \cellcolor{gray!12}0.000 & \cellcolor{gray!12}0.003 & \cellcolor{gray!12}0.000 & \cellcolor{gray!12}0.009 & \cellcolor{gray!12}0.209 & \cellcolor{gray!12}0.368 & \cellcolor{gray!12}0.002 & \cellcolor{gray!12}0.072 & \cellcolor{gray!12}\textbf{0.001} & \cellcolor{gray!12}\textbf{0.083} \\
        \cmidrule(lr){2-13}
         & \multirow{3}{*}{Score $\uparrow$} & RLVR & 0.504 & 0.329 & -0.169 & 0.222 & -0.198 & -0.309 & 0.300 & 0.026 & 0.195 & 0.088 \\
         &  & RLCR & 0.500 & 0.360 & 0.096 & 0.218 & -0.173 & -0.311 & 0.355 & 0.123 & 0.236 & 0.146 \\
         &  & \cellcolor{gray!12}RLEC & \cellcolor{gray!12}0.504 & \cellcolor{gray!12}0.353 & \cellcolor{gray!12}0.091 & \cellcolor{gray!12}0.258 & \cellcolor{gray!12}-0.015 & \cellcolor{gray!12}-0.132 & \cellcolor{gray!12}0.336 & \cellcolor{gray!12}0.112 & \cellcolor{gray!12}\textbf{0.261} & \cellcolor{gray!12}\textbf{0.188} \\
        \bottomrule
    \end{tabular}%
    }
\end{table}

Table~\ref{tab:calibration_results} reports Micro results pooled over all examples and Macro results averaged equally across benchmarks. Micro reflects aggregate performance and is less affected by difficult benchmarks such as SimpleQA-Verified, whereas Macro prevents larger benchmarks from dominating.
RLEC largely preserves the answer ability obtained by RLVR, with comparable Forced Accuracy at both scales. Compared with RLCR, the 4B model achieves lower AOC but also lower Score, indicating a trade-off between target compliance and utility. At 30B, RLEC achieves both lower AOC and higher Score, outperforming RLCR in both respects. These results demonstrate the feasibility of direct training with explicit confidence targets and show that RLEC remains effective at the larger scale.

\subsection{Ablation Studies}
\label{sec:ablation}

To answer \textbf{RQ3}, we study two main design choices in RLEC: omitting the reward standard deviation from group advantage estimation and selecting the stability coefficient \(b\).

\begin{figure}[t]
    \centering
    \includegraphics[width=\textwidth]{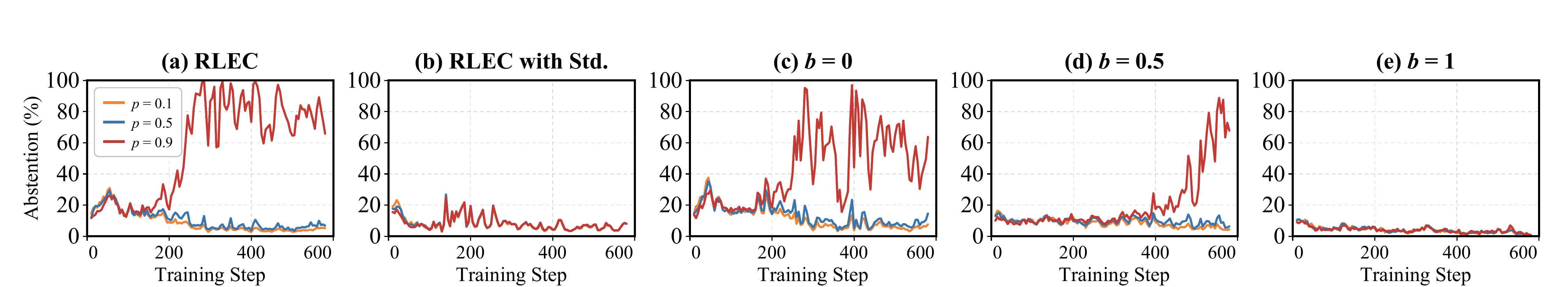}
    \caption{\textbf{Ablation training curves.} Abstention rates at \(p\in\{0.1,0.5,0.9\}\) for (a) RLEC, (b) RLEC with standard deviation normalization, and fixed (c) \(b=0\), (d) \(b=0.5\), and (e) \(b=1\).}
    \label{fig:ablation_curves}
\end{figure}

\paragraph{Standard Deviation.} As shown in Figure~\ref{fig:ablation_curves}(b), adding group standard deviation normalization produces nearly overlapping abstention curves across different \(p\). This supports our analysis that normalization weakens the reward scale signal induced by \(p\). Also, Appendix~\ref{sec:ablation_std} shows that removing the group standard deviation does not consistently improve or degrade answer ability.

\paragraph{Choice of Stability Coefficient.}
Panels (c), (d), and (e) of Figure~\ref{fig:ablation_curves} illustrate the different behaviors induced by fixed choices of \(b\). With \(b=0\), the model distinguishes the confidence targets relatively early, but its abstention rate fluctuates sharply. With \(b=1\), the curves nearly overlap, and the model fails to distinguish the targets. The intermediate choice \(b=0.5\) produces separation only late in training, indicating less efficient learning and potentially less reliable target differentiation.

We also evaluate the different choices on the test benchmarks. Forced Accuracy, AOC, and Score remain broadly consistent, demonstrating that the proposed reward family is robust to the specific choice of \(b\). The scheduled configuration achieves the highest aggregate Forced Accuracy while retaining competitive calibration performance. Because we consider preserving answer ability a primary requirement for behavioral calibration, we select it as our default configuration. Complete results and further discussion are provided in Appendix~\ref{app:detailed_ablation_results}.


\subsection{Additional Properties}
\label{sec:additional_properties}

To answer \textbf{RQ4}, we examine fine-grained target control, the operational implicit confidence profile, and the risk--coverage trade-off.

\subsubsection{Fine-Grained Confidence Control}
\label{sec:fine_grained_control}

\begin{figure}[tb]
    \centering
    \includegraphics[width=\textwidth]{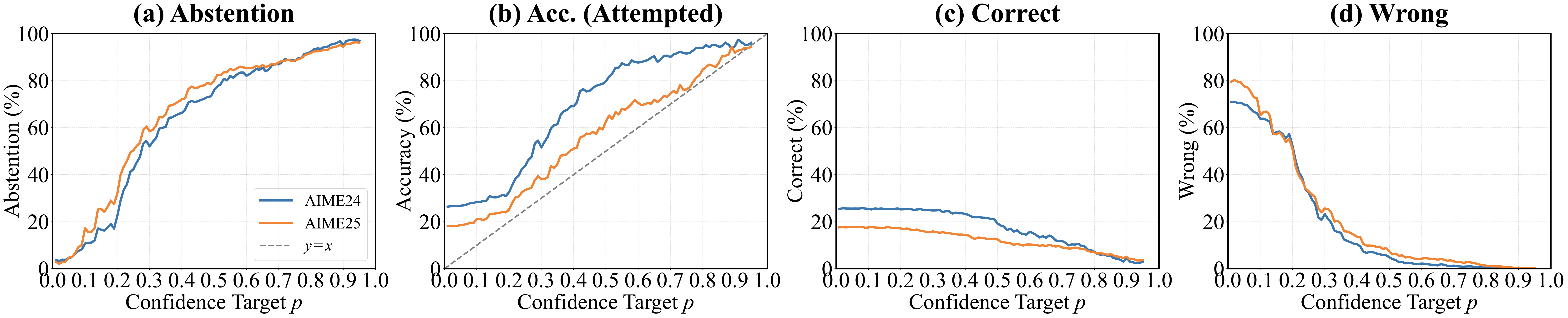}
    \caption{\textbf{Fine-grained confidence control on AIME24 and AIME25.} Targets vary at intervals of \(0.01\) up to \(p=0.95\), with Avg@1024 for each target. (a) Abstention rate. (b) Accuracy among answered questions. (c,d) Proportions of all responses that are correct and wrong. Training uses targets up to \(p=0.90\).}
    \label{fig:fine_grained_control}
\end{figure}

We evaluate RLEC on AIME24 and AIME25 with confidence targets spaced at intervals of \(0.01\), extending up to \(p=0.95\). We choose these benchmarks because their difficulty provides an informative regime for studying fine-grained control: their relatively low accuracy avoids the saturation observed on easier benchmarks such as GSM8K, while a meaningful subset of problems remains within the model's reasoning capability. In contrast, errors on knowledge-intensive benchmarks such as SimpleQA more often reflect missing factual knowledge, leaving less room for confidence control to distinguish solvable from unsolvable examples. Each target is evaluated using Avg@1024. Figure~\ref{fig:fine_grained_control} reports the abstention rate, attempted accuracy, and the proportions of all responses that are correct and wrong. Because training uses only targets up to \(p=0.90\), the interval \(p\in(0.90,0.95]\) additionally evaluates behavior beyond the training range.

As shown in Figure~\ref{fig:fine_grained_control}, the abstention rate rises steadily and attempted accuracy generally improves as \(p\) increases, demonstrating that RLEC responds to fine-grained changes in the target, including values such as \(p=0.01\) and \(p=0.02\). The curves are not strictly monotonic at every adjacent target, which we attribute to finite sample evaluation noise, but their overall trends remain stable. Attempted accuracy also generally meets or exceeds the target.

In contrast, although the RLCR prompt does not restrict the number of decimal places in its confidence output, as shown in Figure~\ref{fig:prompt_templates}, all confidence values produced by the evaluated RLCR model lie on a one-decimal
grid; we do not observe intermediate values such as \(0.12\). This observation
is consistent with prior findings \citep{chen2026out,dai2026rescaling}. Consequently, even if an external system applies arbitrary thresholds to these outputs, it cannot recover additional operating points between adjacent confidence values. The evaluated RLCR model therefore provides substantially coarser control than directly conditioning RLEC on targets at a resolution of \(0.01\).

\paragraph{Selective Error Filtering.}
Panels (c) and (d) further show that the number of wrong answers decreases much faster than the number of correct answers as \(p\) increases. This effect is particularly clear at lower confidence targets, where the proportion of correct answers remains nearly unchanged while many wrong answers are filtered out. RLEC therefore initially improves attempted accuracy primarily by abstaining from likely errors rather than discarding correct answers, yielding a favorable trade-off between coverage and answer reliability.

\paragraph{Extrapolation Beyond the Training Range.}
RLEC is trained only with targets up to \(p=0.90\), whereas Figure~\ref{fig:fine_grained_control} extends the evaluation to the unseen range \(p\in(0.90,0.95]\). Despite small local fluctuations and the reduced coverage at these high targets, all four metrics remain stable and continue their overall trends without an abrupt behavioral change. These results suggest that RLEC remains responsive even to confidence targets beyond the range observed during training.

\subsubsection{Implicit Confidence Profile}
\label{sec:implicit_confidence_profile}

RLEC directly accepts an explicit confidence target \(p\) and decides whether to answer, rather than outputting a confidence score. Nevertheless, to better understand its aggregate behavior and facilitate comparison with RLCR, we construct an operational approximation of its implicit confidence profile. We measure the abstention rate at \(p\in\{0,0.1,\dots,0.9\}\), treat it as an estimate of the cumulative mass below \(p\), and use differences between neighboring targets as interval masses. This construction summarizes the aggregate behavior of the \(p\)-conditioned policies and does not assume a shared sample-level latent confidence variable. For RLCR, we directly use its reported confidence.

\begin{figure}[tb]
    \centering
    \includegraphics[width=\textwidth]{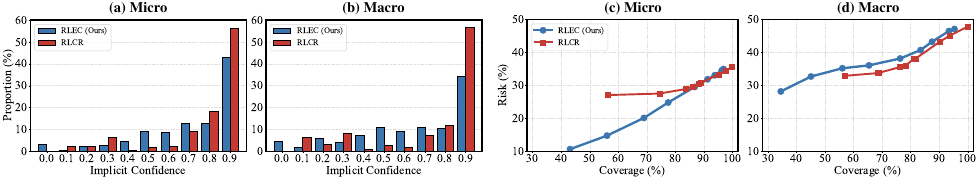}
    \caption{\textbf{Implicit confidence profiles and risk--coverage trade-offs on the 30B model.} (a,b) Micro and Macro profiles, constructed from RLEC's abstention behavior and RLCR's reported confidence. (c,d) Micro and Macro risk--coverage curves. RLEC's operational profile is more dispersed, and its risk--coverage curves extend to lower-coverage operating points.}
    \label{fig:confidence_risk_profiles}
\end{figure}

Figure~\ref{fig:confidence_risk_profiles}(a,b) compares the resulting profiles on the 30B model. Under this construction, RLEC has a higher standard deviation (\(0.2335\) versus \(0.2126\) for Micro and \(0.2713\) versus \(0.2646\) for Macro), with less mass in the highest interval and more in the intermediate intervals. These results suggest that RLEC supports a broader range of aggregate behavior across confidence targets.

\subsubsection{Risk-Coverage Trade-off}
\label{sec:risk_coverage}

Figure~\ref{fig:confidence_risk_profiles}(c,d) plots risk, the error rate among answered questions, against coverage, the fraction of questions answered, on the 30B model. A lower curve is better, since it means fewer errors at the same coverage. Overall, the two methods perform similarly: RLEC is slightly better in the Micro setting, while RLCR is slightly better in the Macro setting. The main difference is in operating range. In both settings, RLEC extends to a lower coverage than RLCR, so it can abstain more and keep risk lower when the target calls for high reliability, whereas RLCR cannot reach the same conservative operating points. This is a potential practical advantage in applications that demand high reliability.

\section{Related Work}
\subsection{Reinforcement Learning with Verifiable Rewards}

Reinforcement learning with verifiable rewards (RLVR)~\cite{deepseekmath,deepseekr1,yu2026dapo,liu2025understanding,chen2025minimax} has become a common approach for improving LLM reasoning. A widely used algorithm is Group Relative Policy Optimization (GRPO)~\cite{deepseekmath}, which estimates advantages from multiple responses to the same prompt by subtracting the group mean and dividing by the group standard deviation. Prior work has examined the benefits and drawbacks of this normalization. Dr.\ GRPO~\cite{liu2025understanding} removes it to avoid weighting questions according to their difficulty. ~\citet{turtel2025outcome} remove standard deviation normalization to preserve the effect of large prediction errors, while ~\citet{bereket2025uncalibrated} show that removing it reduces biased advantages and overconfidence. In contrast, ~\citet{ge2026grpo} view normalization as adaptive gradient scaling that can improve convergence and stability, while CAL-GRPO~\citep{ildiz2026learning} also reports that retaining normalization improves sample efficiency across its GRPO variants.

These studies mainly examine how standard deviation normalization affects optimization and model performance. Unlike all of them, RLEC uses the reward scale to encode the confidence target \(p\). We show that for groups with two distinct reward values and fixed proportions, normalization removes the effect of the reward scale induced by \(p\) on the advantages. Such groups are common during training, helping explain the failure to distinguish confidence targets in prior work~\cite{wu2025mitigating}. RLEC retains this signal through mean centering without standard deviation normalization.

\subsection{Hallucination Mitigation and Behavioral Calibration}

Hallucination~\cite{ji2023survey,huang2025survey} remains a major source of errors in LLMs. In \emph{Why Language Models Hallucinate}, \citet{kalai2025language} argue that common training objectives encourage models to guess rather than admit uncertainty. This has motivated RL methods that reward correct answers, penalize errors, and allow abstention~\cite{wei2025truthrl,mohamadi2025honesty,jha2026rewarding,pan2026tiar}. However, these methods encode a fixed risk preference and cannot adapt their decision to answer or abstain to different user requirements at inference time. \citet{wu2025mitigating} propose Explicit Risk Thresholding (ERT), which extends this setting to confidence targets specified in the prompt, but report that it fails to learn distinct behaviors across targets.

Another line of work trains or prompts models to express confidence~\cite{lin2022teaching,tian2023just}. For example, \citet{damani2026beyond} propose Reinforcement Learning with Calibration Rewards (RLCR), which trains models to produce an answer together with a calibrated confidence score. However, these confidence estimates typically need to be converted into decisions by an external rule or system, and some approaches further require a separate calibrator or calibration set~\cite{kamath2020selective,wang2026safer}. In contrast, RLEC enables models to learn how to directly adapt their answer or abstention decisions to explicit confidence targets, without an external rule for converting confidence estimates into decisions.

\section{Conclusion}
\label{sec:conclusion}

We study explicit confidence targets as a more flexible and direct way to control LLM abstention. We show that the failure of prior work results from the loss of signals and the unstable scale of reward. Based on this analysis, we propose RLEC, which preserves the confidence target signal, introduces a stable reward family, and uses reward scheduling to balance effective learning and training stability. Experiments across diverse benchmarks and model scales show that explicit confidence target training is feasible. RLEC largely preserves general capabilities, achieves strong calibration, and supports fine-grained control without an external decision system. More broadly, RLEC opens a new and direct path toward behavioral calibration with explicit, user-specified confidence targets.

\clearpage
\begingroup
\setlength{\bibsep}{3pt plus 0.5pt}
\bibliographystyle{plainnat}
\bibliography{references}
\endgroup

\clearpage
\beginappendix
\begingroup
\setcounter{tocdepth}{2}
\makeatletter
\@starttoc{atoc}
\let\seed@addcontentsline\addcontentsline
\renewcommand{\addcontentsline}[3]{%
    \seed@addcontentsline{#1}{#2}{#3}%
    \ifstrequal{#1}{toc}{\seed@addcontentsline{atoc}{#2}{#3}}{}%
}
\makeatother
\section{Detailed Experimental Setup}
\label{app:experimental_setup}

\subsection{Implementation Details}
\label{app:implementation_details}

\begin{wraptable}{r}{0.48\textwidth}
    \centering
    \footnotesize
    \setlength{\tabcolsep}{6pt}
    \caption{\textbf{Hyperparameters.}}
    \label{tab:hyperparameters}
    \begin{tabular}{lc}
        \toprule
        Hyperparameter & Value \\
        \midrule
        \multicolumn{2}{l}{\textit{Training}} \\
        \texttt{max\_response\_length} & 4096 \\
        \texttt{train\_batch\_size} & 256 \\
        \texttt{ppo\_mini\_batch\_size} & 128 \\
        \texttt{optim.lr} & \(1\times10^{-6}\) \\
        \texttt{rollout.temperature} & 1.0 \\
        \texttt{rollout.n} & 8 \\
        \midrule
        \multicolumn{2}{l}{\textit{Evaluation}} \\
        \texttt{max\_tokens} & 4096 \\
        \texttt{temperature} & 1.0 \\
        \texttt{top\_p} & 0.95 \\
        \bottomrule
    \end{tabular}
\end{wraptable}

\paragraph{Training and Evaluation.}
We use \texttt{verl} for RL training~\cite{sheng2025hybridflow}, with the hyperparameters for all RL methods summarized in Table~\ref{tab:hyperparameters}. We do not use Clip-Higher, entropy regularization, or a KL penalty. For RLCR, we determine abstention by applying the requested confidence target as an external threshold to its reported confidence. For all other methods, abstentions are identified by matching the designated abstention string. For \(p=0\), we use a binary reward and evaluation criterion: only a correct answer is treated as correct, whereas incorrect answers and abstentions are both treated as incorrect. For AOC, accuracy is set to zero when no answers are attempted, balancing calibration against utility. 
Due to limited computational resources, we do not conduct multiple training runs for each setting. Instead, we report Avg@16 for AIME and Avg@2 for all other benchmarks to reduce evaluation variance.

\paragraph{Prompts.}
The prompt templates used by RLEC and the baselines are shown in Figure~\ref{fig:prompt_templates}. For other abstention methods, including ERT and TruthRL, we use the RLEC prompt during both training and evaluation. For RLVR, we use its standard prompt during training and at \(p=0\) to represent the conventional RL setting and compare answer ability. At \(p>0\), we use the RLEC prompt to demonstrate that standard RLVR training alone does not enable target-conditioned abstention. For SimpleQA-Verified and AA-Omniscience, we additionally use the grading prompt shown in Figure~\ref{fig:grading_prompt} and use Qwen3.6-27B as the judge. Since abstention is identified separately, this grader only determines whether the predicted answer is semantically equivalent to the reference answer.

\begin{figure}[t]
    \centering

    \begin{tcolorbox}[
        width=\textwidth,
        colback=gray!4,
        colframe=black!55,
        boxrule=0.6pt,
        arc=1mm,
        left=8pt,
        right=8pt,
        top=6pt,
        bottom=6pt,
        title={\textbf{RLEC Prompt}},
        fonttitle=\bfseries
    ]
    \small\rmfamily\raggedright
    Work through this problem step by step, then honestly assess how confident you are that your solution is correct. Only give a final answer if you are at least \texttt{\{p\}} confident that it is correct; otherwise do not guess and answer with \texttt{\textbackslash boxed\{I am not sure.\}} instead. Put your final answer inside \texttt{\textbackslash boxed\{\}}. If you are not confident enough, put \texttt{\textbackslash boxed\{I am not sure.\}} instead.
    \end{tcolorbox}

    \vspace{4pt}

    \begin{tcolorbox}[
        width=\textwidth,
        colback=gray!4,
        colframe=black!55,
        boxrule=0.6pt,
        arc=1mm,
        left=8pt,
        right=8pt,
        top=6pt,
        bottom=6pt,
        title={\textbf{RLCR Prompt}},
        fonttitle=\bfseries
    ]
    \small\rmfamily\raggedright
    Work through this problem step by step, then honestly assess how confident you are that your solution is correct. Put your final answer inside \texttt{\textbackslash boxed\{\}}, then output your confidence that the answer is correct as a number between 0 and 1 (inclusive) enclosed within \texttt{<confidence> </confidence>} tags.
    \end{tcolorbox}

    \vspace{4pt}

    \begin{tcolorbox}[
        width=\textwidth,
        colback=gray!4,
        colframe=black!55,
        boxrule=0.6pt,
        arc=1mm,
        left=8pt,
        right=8pt,
        top=6pt,
        bottom=6pt,
        title={\textbf{RLVR Prompt}},
        fonttitle=\bfseries
    ]
    \small\rmfamily\raggedright
    Work through this problem step by step. Put your final answer inside \texttt{\textbackslash boxed\{\}}.
    \end{tcolorbox}

    \caption{\textbf{Prompt templates used by RLEC and the baselines.}}
    \label{fig:prompt_templates}
\end{figure}

\begin{figure}[t]
    \centering
    \begin{tcolorbox}[
        width=\textwidth,
        colback=gray!4,
        colframe=black!55,
        boxrule=0.6pt,
        arc=1mm,
        left=8pt,
        right=8pt,
        top=6pt,
        bottom=6pt,
        title={\textbf{Correctness-Grading System Prompt}},
        fonttitle=\bfseries
    ]
\small\rmfamily\raggedright
You are a grading assistant. Decide whether the predicted answer is semantically equivalent to the reference gold answer for the given question. Ignore formatting differences, units notation, LaTeX wrappers, and trivial rephrasing. Reply with a single token: \texttt{YES} or \texttt{NO}.
    \end{tcolorbox}
    \caption{\textbf{System prompt used to grade SimpleQA-Verified and AA-Omniscience.}}
    \label{fig:grading_prompt}
\end{figure}

\subsection{Benchmarks}
\label{app:benchmark_details}

We use the standard versions of the evaluation benchmarks except for the following two settings.

\paragraph{HotpotQA.} We use the modified version~\cite{damani2026beyond} of HotpotQA~\cite{yang2018hotpotqa}. Each example contains eight paragraphs, while zero, one, or both of the two relevant paragraphs are removed to test whether the model abstains when the available evidence is incomplete.

\paragraph{MMLU-Pro.} We evaluate the four science subsets in MMLU-Pro~\cite{wang2024mmlu}: biology, chemistry, physics, and health. Since the full benchmark contains substantially more examples than the other benchmarks, this selection prevents MMLU-Pro from dominating the aggregate results.

\subsection{Baselines}
\label{app:baseline_details}

\paragraph{RLVR.} RLVR uses a binary correctness reward and does not explicitly reward abstention:
\begin{equation}
R_{\mathrm{RLVR}}=
\begin{cases}
1, & \text{correct},\\
0, & \text{otherwise}.
\end{cases}
\end{equation}
In the main experiments, we remove its group standard deviation normalization to align its advantage with RLEC and enable a closer comparison. Appendix~\ref{sec:ablation_std} examines the effect of this choice.

\paragraph{TruthRL.}
TruthRL~\cite{wei2025truthrl} uses a fixed ternary reward that penalizes incorrect answers while assigning zero reward to abstentions:
\begin{equation}
R_{\mathrm{TruthRL}}=
\begin{cases}
1, & \text{correct},\\
-1, & \text{incorrect},\\
0, & \text{abstention}.
\end{cases}
\end{equation}
Our framework is compatible with TruthRL by setting \(p=0.5\), \(b=0\), and retaining group standard-deviation normalization.

\paragraph{ERT.}
ERT~\cite{wu2025mitigating} samples a confidence target \(p\) and applies its corresponding error cost during standard GRPO training:
\begin{equation}
R_{\mathrm{ERT}}(p)=
\begin{cases}
1, & \text{correct},\\
-\dfrac{p}{1-p}, & \text{incorrect},\\
0, & \text{abstention}.
\end{cases}
\end{equation}
Our framework is compatible with ERT by setting \(b=0\) and retaining group standard-deviation normalization.

\paragraph{RLCR.} RLCR~\cite{damani2026beyond} trains the model to produce an answer together with a confidence score \(c\in[0,1]\). Let \(y=1\) if the answer is correct and \(y=0\) otherwise; its reward combines correctness with the Brier score:
\begin{equation}
R_{\mathrm{RLCR}}(y,c)=y-(c-y)^2.
\end{equation}
For RLCR, we remove group standard deviation normalization to obtain better calibration~\cite{turtel2025outcome,bereket2025uncalibrated}. At evaluation time, we use an external decision system that retains the answer when \(c\geq p\) and replaces it with an abstention otherwise.

\section{Additional Experimental Results}
\label{app:additional_results}

\subsection{Choice of Stability Coefficient}
\label{app:detailed_ablation_results}

\begin{table}[t]
    \centering
    \caption{Forced answer accuracy, average overconfidence (AOC), and average score for RLVR and different stability coefficient and scheduling choices. The best Micro and Macro results among variants that distinguish confidence targets are shown in bold; RLVR and $b=1$ are excluded from this comparison because they cannot distinguish among targets.}
    \label{tab:ablation_full}
    \resizebox{\textwidth}{!}{%
    \begin{tabular}{@{}llcccccccccc@{}}
        \toprule
        & & \multicolumn{3}{c}{Mathematical Reasoning} & \multicolumn{3}{c}{Knowledge \& Hallucination} & \multicolumn{2}{c}{Scientific Reasoning} & \multicolumn{2}{c}{Avg.} \\
        \cmidrule(lr){3-5}\cmidrule(lr){6-8}\cmidrule(lr){9-10}\cmidrule(lr){11-12}
        Metric & Method & GSM8K & MATH & AIME & Hotpot & OmniSci & SimpleQA & MMLU-Pro & GPQA & Micro & Macro \\
        \midrule
        \multirow{7}{*}{Acc. $\uparrow$} & RLVR & 93.33 & 69.70 & 20.83 & 69.50 & 13.75 & 5.20 & 68.47 & 35.91 & 58.18 & 47.09 \\
         & \cellcolor{gray!12}Ours & \cellcolor{gray!12}92.15 & \cellcolor{gray!12}70.50 & \cellcolor{gray!12}19.79 & \cellcolor{gray!12}67.60 & \cellcolor{gray!12}14.17 & \cellcolor{gray!12}4.75 & \cellcolor{gray!12}67.37 & \cellcolor{gray!12}36.24 & \cellcolor{gray!12}\textbf{57.40} & \cellcolor{gray!12}\textbf{46.57} \\
         & $b=0$ & 91.09 & 69.30 & 16.25 & 66.00 & 14.25 & 4.65 & 67.32 & 36.35 & 56.95 & 45.65 \\
         & $b=0.5$ & 92.46 & 70.90 & 15.62 & 67.70 & 14.75 & 4.25 & 67.40 & 33.00 & 57.18 & 45.76 \\
         & $b=1$ & 91.81 & 70.40 & 19.38 & 67.80 & 14.08 & 5.30 & 69.49 & 34.56 & 58.29 & 46.60 \\
         & LR Schedule & 91.77 & 69.80 & 16.46 & 64.00 & 14.08 & 4.10 & 67.07 & 34.68 & 56.69 & 45.25 \\
         & Sine Schedule & 93.03 & 69.60 & 20.00 & 67.80 & 13.75 & 4.45 & 65.77 & 35.12 & 56.63 & 46.19 \\
        \midrule
        \multirow{7}{*}{AOC $\downarrow$} & RLVR & 0.000 & 0.027 & 0.258 & 0.033 & 0.292 & 0.389 & 0.032 & 0.175 & 0.049 & 0.151 \\
         & \cellcolor{gray!12}Ours & \cellcolor{gray!12}0.000 & \cellcolor{gray!12}0.000 & \cellcolor{gray!12}0.090 & \cellcolor{gray!12}0.010 & \cellcolor{gray!12}0.259 & \cellcolor{gray!12}0.381 & \cellcolor{gray!12}0.015 & \cellcolor{gray!12}0.157 & \cellcolor{gray!12}0.013 & \cellcolor{gray!12}0.114 \\
         & $b=0$ & 0.000 & 0.002 & 0.037 & 0.008 & 0.178 & 0.388 & 0.016 & 0.144 & 0.010 & 0.096 \\
         & $b=0.5$ & 0.000 & 0.001 & 0.018 & 0.000 & 0.207 & 0.370 & 0.015 & 0.141 & \textbf{0.008} & \textbf{0.094} \\
         & $b=1$ & 0.000 & 0.010 & 0.020 & 0.031 & 0.251 & 0.384 & 0.025 & 0.154 & 0.029 & 0.109 \\
         & LR Schedule & 0.000 & 0.004 & 0.013 & 0.017 & 0.227 & 0.382 & 0.013 & 0.144 & 0.011 & 0.100 \\
         & Sine Schedule & 0.000 & 0.003 & 0.009 & 0.016 & 0.293 & 0.399 & 0.015 & 0.148 & 0.022 & 0.111 \\
        \midrule
        \multirow{7}{*}{Score $\uparrow$} & RLVR & 0.480 & 0.261 & -0.162 & 0.237 & -0.173 & -0.213 & 0.237 & -0.090 & 0.167 & 0.072 \\
         & \cellcolor{gray!12}Ours & \cellcolor{gray!12}0.477 & \cellcolor{gray!12}0.333 & \cellcolor{gray!12}0.056 & \cellcolor{gray!12}0.264 & \cellcolor{gray!12}-0.048 & \cellcolor{gray!12}-0.107 & \cellcolor{gray!12}0.262 & \cellcolor{gray!12}-0.016 & \cellcolor{gray!12}0.214 & \cellcolor{gray!12}0.153 \\
         & $b=0$ & 0.474 & 0.304 & 0.054 & 0.267 & -0.010 & -0.063 & 0.264 & -0.001 & \textbf{0.222} & 0.161 \\
         & $b=0.5$ & 0.477 & 0.334 & 0.048 & 0.276 & -0.025 & -0.061 & 0.258 & 0.015 & \textbf{0.222} & \textbf{0.165} \\
         & $b=1$ & 0.474 & 0.305 & 0.073 & 0.241 & -0.087 & -0.129 & 0.247 & -0.034 & 0.198 & 0.136 \\
         & LR Schedule & 0.473 & 0.316 & 0.063 & 0.241 & -0.011 & -0.066 & 0.261 & 0.001 & 0.219 & 0.160 \\
         & Sine Schedule & 0.478 & 0.330 & 0.074 & 0.257 & -0.101 & -0.180 & 0.264 & -0.007 & 0.204 & 0.139 \\
        \bottomrule
    \end{tabular}%
    }
\end{table}

\begin{figure}[t]
    \centering
    \includegraphics[width=\textwidth]{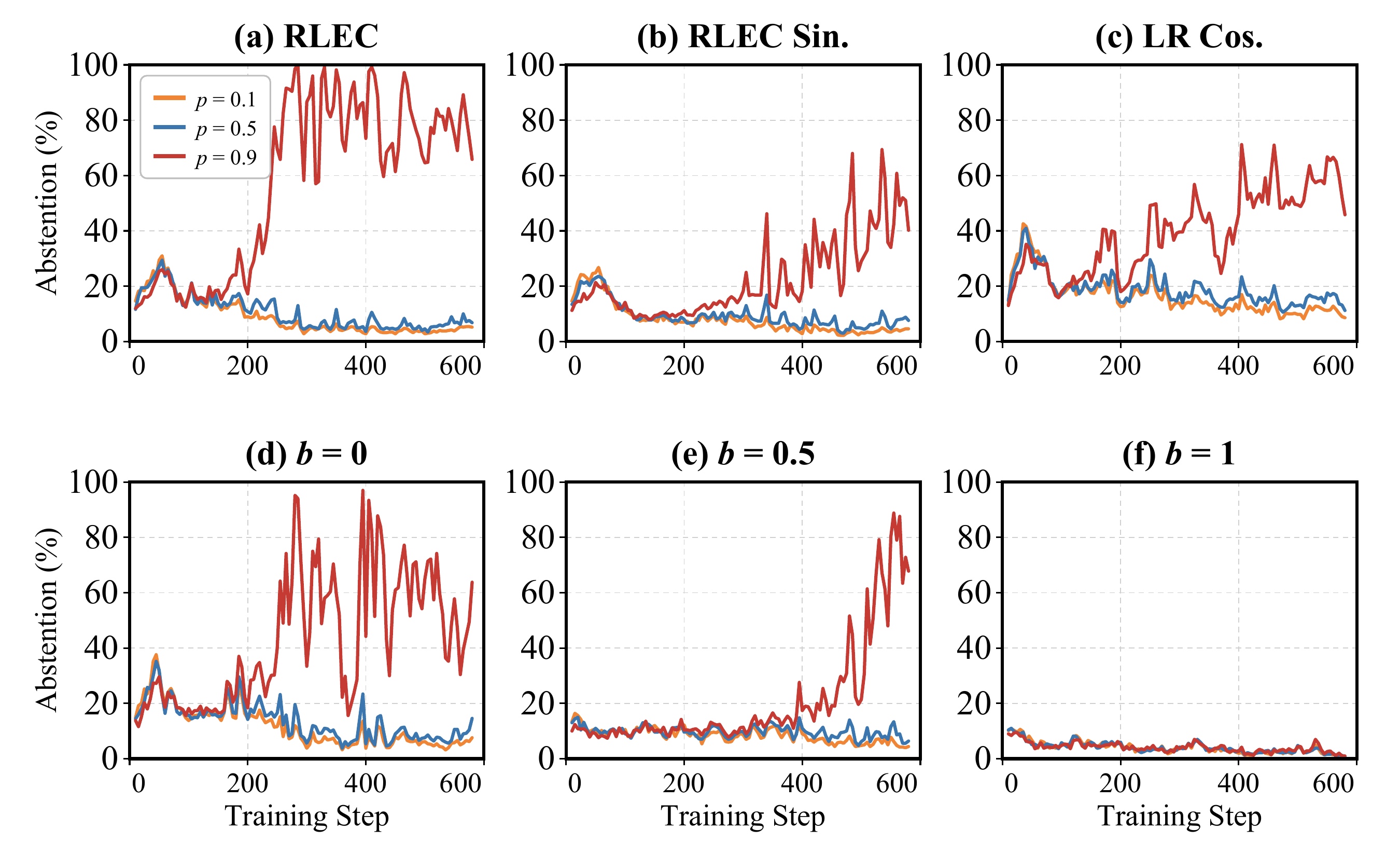}
    \caption{Training dynamics of the abstention rate on HotpotQA under different choices of the stability coefficient and scheduling strategy. RLEC uses the linear schedule of \(b\), while RLEC Sin. uses a sinusoidal schedule. Curves correspond to target confidence levels \(p\in\{0.1,0.5,0.9\}\).}
    \label{fig:training_curves}
\end{figure}
We examine the choice of \(b\) within our proposed reward family, considering fixed values as well as linear and sinusoidal schedules. As shown in Equation~\ref{eq:stable_reward_expectation}, these choices preserve the same theoretical decision rule but induce different balances among signal strength, optimization stability, and answer ability. Table~\ref{tab:ablation_full} reports their Forced Accuracy, AOC, and Score, reflecting answer ability, target compliance, and utility, respectively. We additionally include LR Schedule as a possible variant, which applies a global cosine learning-rate decay.

\paragraph{Forced Accuracy.} Among the configurations that distinguish confidence targets, Forced Accuracy remains broadly stable, with the linear schedule obtaining the highest aggregate Forced Accuracy. The clearest differences appear on AIME, where fixed \(b=0\), fixed \(b=0.5\), and LR Schedule are all more than three points below the linear schedule. Because AIME is particularly difficult and benefits substantially from reinforcement learning, this result is consistent with training stability being especially important for preserving abilities acquired during this stage. Differences on most other benchmarks are smaller or less consistent, particularly on knowledge benchmarks such as SimpleQA-Verified and AA-Omniscience, where performance largely depends on knowledge acquired during pretraining. LR Schedule performs worse in aggregate Forced Accuracy, indicating that scheduling the relative reward scale across confidence targets is not equivalent to applying a global learning-rate decay.

\paragraph{Calibration Performance.}
All choices of \(b\) substantially improve calibration over RLVR, while the differences among these choices remain modest. In particular, \(b=0.5\) obtains the lowest aggregate AOC and the highest Macro Score, while retaining relatively strong Forced Accuracy.

\paragraph{Training Dynamics.}
The training dynamics in Figure~\ref{fig:training_curves}, however, reveal an additional distinction that is not captured by the final aggregate results. Under the linear schedule, the behavior for \(p=0.9\) separates from the lower-confidence targets around the middle of training and remains distinctly ordered thereafter. Fixed \(b=0\) also distinguishes the targets relatively early, but exhibits larger fluctuations. With fixed \(b=0.5\), clear separation emerges only after approximately 400 steps despite its favorable final calibration and relatively strong Forced Accuracy, suggesting that it may not establish target conditioned behavior as reliably throughout optimization and may be more sensitive to the available training horizon. With \(b=1\), the curves remain nearly indistinguishable, consistent with the absence of a target-dependent reward gap. Overall, smaller values of \(b\) strengthen target-dependent learning but reduce stability, whereas larger values stabilize updates but weaken differentiation. The linear schedule balances these effects by establishing the distinctions early and stabilizing them later.

\paragraph{Final Choice.}
We treat preserving answer ability as a primary requirement for behavioral calibration, since calibration improvements are less desirable when accompanied by degradation in the model's underlying answer accuracy. Although the sinusoidal schedule achieves comparable Macro results across all three metrics, it provides no consistent advantage over the linear schedule. We therefore adopt the linear schedule as our default configuration and use it to scale RLEC to the 30B model, because it achieves stronger aggregate Forced Accuracy, establishes earlier and more persistent target separation, and retains a simpler form than the sinusoidal alternative.

\begin{table}[t]
    \centering
    \caption{Forced answer accuracy for RLVR with and without group standard deviation normalization. The best result in each column is shown in bold.}
    \label{tab:rlvr_std_accuracy}
    \resizebox{\textwidth}{!}{%
    \begin{tabular}{@{}lcccccccccc@{}}
        \toprule
        & \multicolumn{3}{c}{Mathematical Reasoning} & \multicolumn{3}{c}{Knowledge \& Hallucination} & \multicolumn{2}{c}{Scientific Reasoning} & \multicolumn{2}{c}{Avg.} \\
        \cmidrule(lr){2-4}\cmidrule(lr){5-7}\cmidrule(lr){8-9}\cmidrule(lr){10-11}
        Method & GSM8K & MATH & AIME & Hotpot & OmniSci & SimpleQA & MMLU-Pro & GPQA & Micro & Macro \\
        \midrule
        RLVR w/o Std. & \textbf{93.33} & 69.70 & \textbf{20.83} & \textbf{69.50} & 13.75 & \textbf{5.20} & 68.47 & 35.91 & 58.18 & 47.09 \\
        RLVR w/ Std. & 92.57 & \textbf{72.30} & 19.17 & 68.30 & \textbf{14.83} & 5.15 & \textbf{68.97} & \textbf{36.80} & \textbf{58.45} & \textbf{47.26} \\
        \bottomrule
    \end{tabular}%
    }
\end{table}

\subsection{Standard Deviation}
\label{sec:ablation_std}
RLEC removes the group standard deviation from advantage estimation to preserve the reward signal. To enable a closer comparison, we also remove this normalization from RLVR in our main experiments. To further isolate its effect on RLVR's answer ability, we compare RLVR with centered advantages against standard RLVR under the same binary reward. As shown in Table~\ref{tab:rlvr_std_accuracy}, their aggregate Forced Accuracy is nearly identical, and neither variant consistently outperforms the other across benchmarks. Thus, removing the group standard deviation provides no clear advantage or disadvantage in answer ability in our experiment setting.

\subsection{Increasing the Rollout Group Size}
\label{app:rollout_group_size}

We further examine whether increasing the rollout group size can help ERT preserve the confidence target signal. We retain the original reward and standard deviation normalization, so this setting differs from ERT only by increasing the number of responses per prompt from \(8\) to \(16\). This change approximately doubles the rollout generation cost.

\begin{figure}[t]
    \centering
    \includegraphics[width=\textwidth]{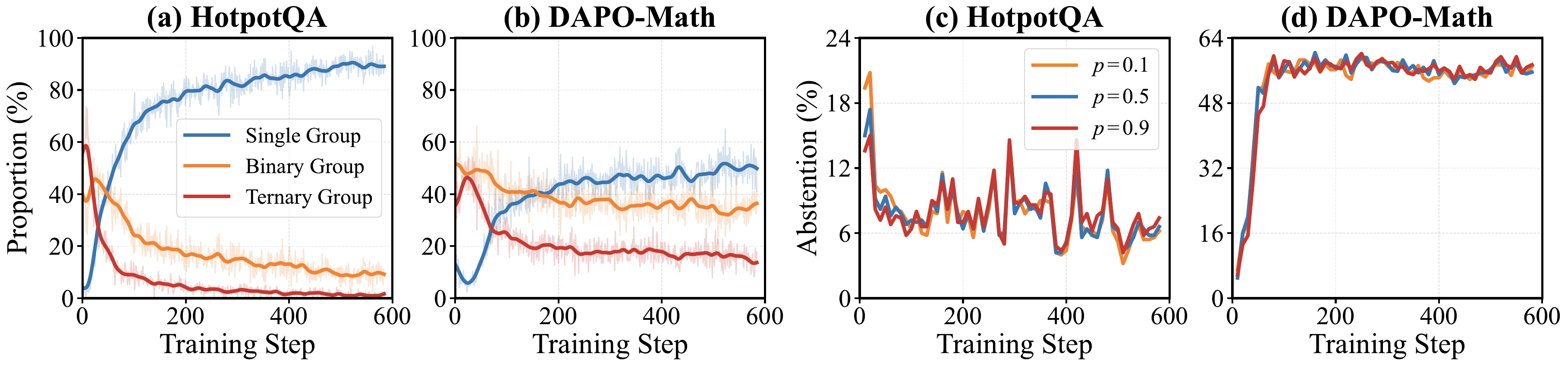}
\caption{\textbf{Scaling ERT to a rollout group size of 16.} (a,b): group compositions during training. (c,d): abstention rates under different confidence targets.}
    \label{fig:ert_rollout16}
\end{figure}

As shown in Figure~\ref{fig:ert_rollout16}, increasing the rollout group size does not enable ERT to distinguish confidence targets. On HotpotQA, groups containing all three reward values rapidly disappear, and groups containing a single reward dominate later training. DAPO Math retains more groups containing all three rewards, but its abstention curves still largely overlap across different values of \(p\). Thus, although a larger rollout group can increase the chance of observing all three outcomes, it does not resolve the loss of target information in groups containing only one or two reward values. These results indicate that increasing rollout size alone cannot replace preserving the target signal during advantage estimation.

\subsection{Scaling RLEC to Longer Contexts}
\label{app:long_context}
Due to computational constraints, our main experiments primarily use a 4K context window for both training and inference. However, difficult reasoning tasks such as AIME may benefit from a larger token budget. To examine whether RLEC can be extended to longer contexts, we increase both the training and inference context windows to 16K while keeping all other hyperparameters identical to those in Table~\ref{tab:hyperparameters}.

\begin{figure}[t]
    \centering
    \includegraphics[width=\textwidth]{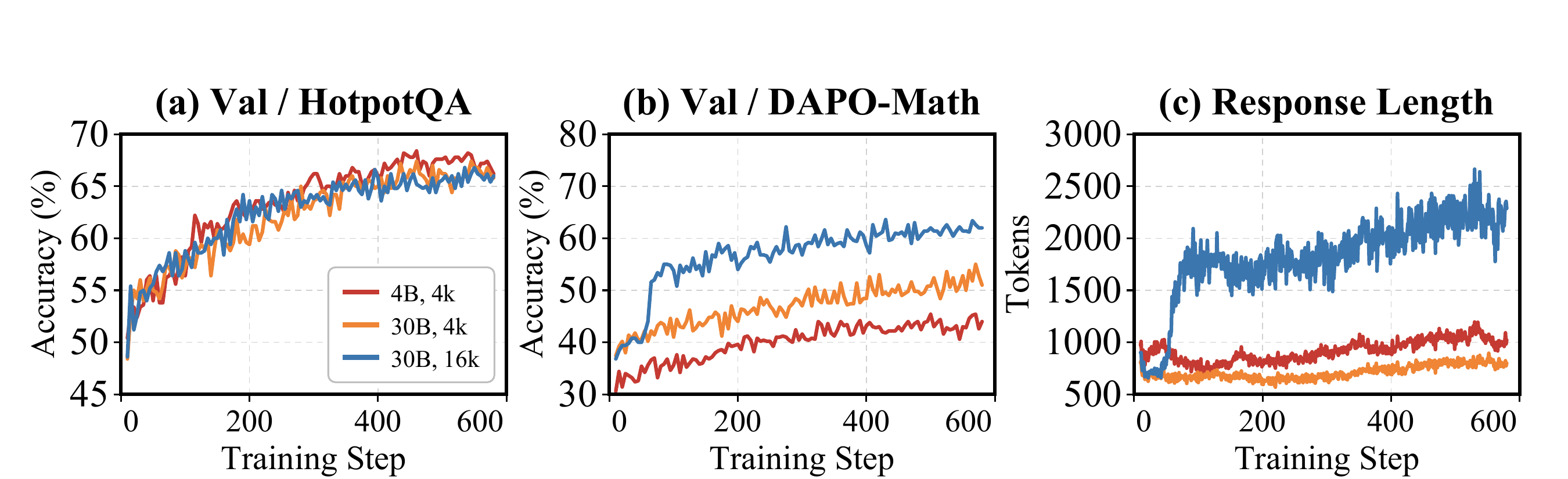}
    \caption{\textbf{Training dynamics under different model scales and context lengths.} We report validation accuracy on HotpotQA and DAPO-Math and the average response length during training.}
    \label{fig:long_context_training}
\end{figure}

\begin{figure}[t]
    \centering
    \includegraphics[width=\textwidth]{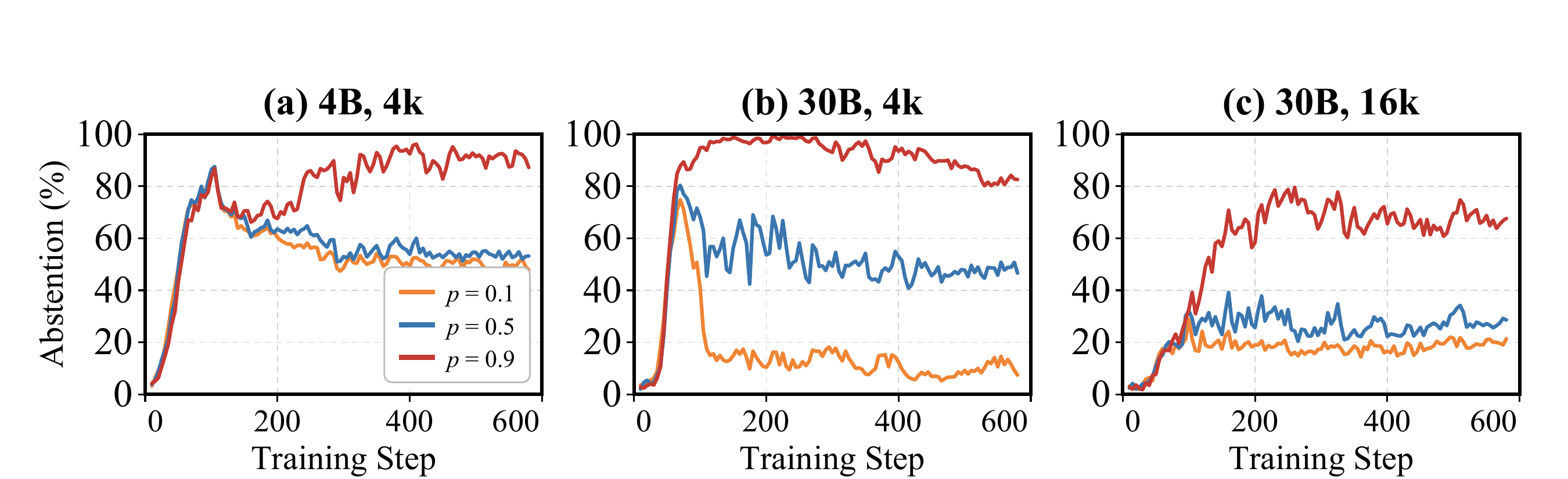}
    \caption{\textbf{Abstention dynamics on DAPO-Math for different model scales and context lengths.}}
    \label{fig:long_context_abstention}
\end{figure}

\begin{table}[t]
    \centering
    \caption{Forced answer accuracy, average overconfidence (AOC), and average score for RLEC-30B with 4K and 16K contexts. The best Micro and Macro results for each metric are shown in bold.}
    \label{tab:context_length}
    \resizebox{\textwidth}{!}{%
    \begin{tabular}{@{}llcccccccccc@{}}
        \toprule
        & & \multicolumn{3}{c}{Mathematical Reasoning} & \multicolumn{3}{c}{Knowledge \& Hallucination} & \multicolumn{2}{c}{Scientific Reasoning} & \multicolumn{2}{c}{Avg.} \\
        \cmidrule(lr){3-5}\cmidrule(lr){6-8}\cmidrule(lr){9-10}\cmidrule(lr){11-12}
        Metric & Context & GSM8K & MATH & AIME & Hotpot & OmniSci & SimpleQA & MMLU-Pro & GPQA & Micro & Macro \\
        \midrule
        \multirow{2}{*}{Acc. $\uparrow$} & 4K & 94.77 & 78.60 & 27.29 & 66.60 & 20.00 & 8.30 & 76.73 & 47.09 & 64.14 & 52.42 \\
         & 16K & 94.28 & 75.00 & 38.96 & 65.60 & 20.08 & 7.55 & 77.99 & 51.68 & \textbf{64.86} & \textbf{53.89} \\
        \midrule
        \multirow{2}{*}{AOC $\downarrow$} & 4K & 0.000 & 0.003 & 0.000 & 0.009 & 0.209 & 0.368 & 0.002 & 0.072 & \textbf{0.001} & 0.083 \\
         & 16K & 0.000 & 0.005 & 0.000 & 0.013 & 0.182 & 0.358 & 0.004 & 0.071 & 0.004 & \textbf{0.079} \\
        \midrule
        \multirow{2}{*}{Score $\uparrow$} & 4K & 0.504 & 0.353 & 0.091 & 0.258 & -0.015 & -0.132 & 0.336 & 0.112 & 0.261 & 0.188 \\
         & 16K & 0.502 & 0.329 & 0.157 & 0.251 & 0.003 & -0.084 & 0.348 & 0.111 & \textbf{0.272} & \textbf{0.202} \\
        \bottomrule
    \end{tabular}%
    }
\end{table}

\paragraph{Training Dynamics.}
As shown in Figure~\ref{fig:long_context_training}, increasing the context window from 4K to 16K leads to substantially longer responses from the 30B model. This additional budget produces a clear improvement in validation reward on DAPO-Math, whereas the HotpotQA trajectories remain largely unchanged. These results suggest that the longer context primarily benefits reasoning-intensive mathematical problems rather than uniformly improving all tasks. The stronger reasoning performance also changes the model's abstention behavior on DAPO-Math. As shown in Figure~\ref{fig:long_context_abstention}, the 16K model abstains less than its 4K counterpart at higher confidence targets, particularly at \(p=0.9\), while preserving the ordering across targets. This behavior is consistent with the stronger model answering more questions even under stringent confidence requirements.

\paragraph{Evaluation Results.}
Table~\ref{tab:context_length} reports the complete evaluation results. As expected, the largest improvement appears on AIME, where Forced Accuracy increases from \(27.29\%\) with the 4K context to \(38.96\%\) with the 16K context. Results on the remaining benchmarks are mixed and generally show smaller differences. Consequently, aggregate Forced Accuracy improves slightly, largely driven by the gain on AIME, while AOC and Score remain broadly comparable. Overall, the longer context primarily benefits difficult mathematical reasoning without materially affecting behavior calibration.

\section{Relation Between Score and the RLCR Reward}
\label{app:score_equivalence}

We show that the score used in our evaluation directly corresponds to the RLCR reward. Let \(y\in\{0,1\}\) indicate whether an answer is correct, and let \(c\in[0,1]\) be the confidence reported by RLCR. Given a confidence target \(p\), the external decision system answers when \(c\geq p\) and abstains otherwise. Its decision can therefore be written as
\begin{equation}
    z_p=\mathbf{1}\{c\geq p\}.
\end{equation}
The score at target \(p\) is
\begin{equation}
    s_p
    =
    z_p(y-p)
    =
    \mathbf{1}\{c\geq p\}(y-p).
\end{equation}
Integrating this score over all confidence targets gives
\begin{align}
    \int_0^1 s_p\,dp
    &=
    \int_0^1 \mathbf{1}\{c\geq p\}(y-p)\,dp \\
    &=
    \int_0^c (y-p)\,dp \\
    &=
    cy-\frac{c^2}{2}.
    \label{eq:integrated_score_appendix}
\end{align}
RLCR uses the following correctness and calibration reward:
\begin{equation}
    R_{\mathrm{RLCR}}(c,y)
    =
    y-(c-y)^2.
\end{equation}
Because \(y\in\{0,1\}\), we have \(y^2=y\). The reward can therefore be rewritten as
\begin{align}
    R_{\mathrm{RLCR}}(c,y)
    &=
    y-c^2+2cy-y^2 \\
    &=
    2cy-c^2 \\
    &=
    2\left(cy-\frac{c^2}{2}\right) \\
    &=
    2\int_0^1 s_p\,dp.
    \label{eq:rlcr_score_equivalence_appendix}
\end{align}
Thus, when the confidence reported by RLCR is converted into an answer or abstention through thresholding, twice the score integrated over all confidence targets is exactly the RLCR reward.

In our experiments, we evaluate a finite set of confidence targets \(\mathcal{P}\) and compute
\begin{equation}
    \widehat{\mathrm{Score}}
    =
    \frac{1}{|\mathcal{P}|}
    \sum_{p\in\mathcal{P}}s_p.
\end{equation}
For uniformly spaced targets over \([0,1)\), this is a Riemann approximation to the integral in Equation~\ref{eq:integrated_score_appendix}. The approximation approaches the exact relation in Equation~\ref{eq:rlcr_score_equivalence_appendix} as the interval between evaluated targets decreases.

The same score also directly corresponds to the RLEC reward. For a given \(p\),
\begin{equation}
    z_p(y-p)
    =
    \begin{cases}
        1-p, & \text{correct answer},\\
        -p, & \text{incorrect answer},\\
        0, & \text{abstention},
    \end{cases}
\end{equation}
which is exactly \(R_p^{(1)}\), the RLEC reward with \(b=1\). Therefore, Score provides a common measure for comparing RLEC and RLCR under explicit confidence targets.

\section{Limitations and Future Work}

Under idealized assumptions, our reward objective induces the decision rule \(q(x)>p\), guaranteeing \(\mathrm{Acc}(p)\geq p\) among answered questions. In practice, limited model capacity, imperfect estimates of answer correctness, finite rollouts, and stochastic optimization can prevent the learned policy from fully realizing this rule, while distribution shifts can further weaken target compliance. Our experiments show clear adaptation to explicit confidence targets and improved aggregate target compliance relative to standard RLVR. However, the requested accuracy is not consistently attained on every benchmark: the remaining gaps are most visible on challenging knowledge-intensive tasks such as SimpleQA-Verified and AA-Omniscience, particularly with the 4B model. Thus, the theoretical objective does not guarantee that the trained model will satisfy every requested target in practice. Future work should explore improved training objectives and data to further improve calibration and reduce target violations, especially on difficult tasks and under distribution shifts, while preserving answer ability and fine-grained target control.

\endgroup

\end{document}